\documentclass[preprint,12pt]{elsarticle}
    
\usepackage{epstopdf}
\usepackage{rotating}
\usepackage{graphicx}
\usepackage{subfig}
\usepackage{amsfonts}
\usepackage{multirow}
\usepackage{enumerate}
\usepackage{comment}
\usepackage{float}
\usepackage{footnote}
\usepackage{hyperref}
\usepackage{stfloats}
\usepackage{algorithm}
\usepackage{algpseudocode}
\usepackage{booktabs}
\usepackage{lscape}
\usepackage{rotating}

\usepackage[cmex10]{amsmath}

\newcommand{\RCr}{\operatorname{\textit{RCr}}}
\newcommand{\pOR}{\operatorname{\textit{pOR}}}

\newcommand{\NMI}{\operatorname{\textit{NMI}}}
\newcommand{\NMP}{\operatorname{\textit{NMP}}}

\makeatletter
\def\ps@pprintTitle{%
 \let\@oddhead\@empty
 \let\@evenhead\@empty
 \def\@oddfoot{}%
 \let\@evenfoot\@oddfoot}
\makeatother

\floatstyle{ruled}
\newfloat{algorithm}{tbp}{loa}
\providecommand{\algorithmname}{Algorithm}
\floatname{algorithm}{\protect\algorithmname}

\usepackage{lscape}
\usepackage{framed}
\usepackage{multirow}

\usepackage{latexsym}
\usepackage[table]{xcolor}

\usepackage{url}
\usepackage{xcolor}
\definecolor{newcolor}{rgb}{.8,.349,.1}

\usepackage{lineno,hyperref}
\modulolinenumbers[5]

\begin{document}

\begin{frontmatter}

\title{Fuzzy $k$-Nearest Neighbors with monotonicity constraints: Moving towards the robustness of monotonic noise}






\author[ugranada]{Sergio~Gonz\'alez}
\ead{sergiogvz@decsai.ugr.es}
\author[ugranada]{Salvador~Garc\'ia}
\ead{salvagl@decsai.ugr.es}
\author[ncku,fin]{Sheng-Tun~Li}
\ead{stli@mail.ncku.edu.tw}
\author[asap]{Robert~John}
\ead{robert.john@nottingham.ac.uk}
\author[ugranada,ukingabdulaziz]{Francisco~Herrera}
\ead{herrera@decsai.ugr.es}

\cortext[cor]{Corresponding author: Sergio~Gonz\'alez}

\address[ugranada]{Department of Computer Science and Artificial Intelligence, University of Granada, 18071 Granada, Spain}
\address[ncku]{Department of Industrial and Information Management and the Institute of Information Management, National Cheng Kung University, Tainan 701, Taiwan}
\address[fin]{Center for Innovative FinTech Business Models, National Cheng Kung University, Tainan 701, Taiwan}
\address[asap]{ASAP Research Group School of Computer Science University of Nottingham NG8 1BB, Nottingham, UK}
\address[ukingabdulaziz]{Faculty of Computing and Information Technology, King Abdulaziz University, Jeddah, Saudi Arabia}

\begin{abstract}

This paper proposes a new model based on Fuzzy $k$-Nearest Neighbors for classification with monotonic constraints, Monotonic Fuzzy $k$-NN (MonF$k$NN). Real-life data-sets often do not comply with monotonic constraints due to class noise. MonF$k$NN incorporates a new calculation of fuzzy memberships, which increases robustness against monotonic noise without the need for relabeling. Our proposal has been designed to be adaptable to the different needs of the problem being tackled. In several experimental studies, we show significant improvements in accuracy while matching the best degree of monotonicity obtained by comparable methods. We also show that MonF$k$NN empirically achieves improved performance compared with Monotonic $k$-NN in the presence of large amounts of class noise.

\end{abstract}

\begin{keyword}
Fuzzy k-NN \sep monotonic constraints \sep ordinal classification \sep ordinal regression \sep class noise.
\end{keyword}

\end{frontmatter}

\section{Introduction}



Monotonic constraints are prior-knowledge of some ordinal classification or regression problems about the order relationships between attributes and class labels \cite{cano19}. Consider the example of house pricing. The following constraints are applied: A bigger house in the same neighborhood is constrained by higher prices as compared to smaller houses with the same features. That is, the classifier decisions should not decrease in the presence of better features while the rest remains the same. These prior constraints are required by many real-life evaluation problems, such as credit risk modeling \cite{chen14} and lecturer evaluation \cite{cano17pro}. These problems are known as Classification with Monotonic Constraints or Monotonic Classification \cite{ben-david95}.

These learning tasks have additional objectives besides accurate models, such as the monotonic consistency of predictions and minimization of the misclassification costs. The latter is also relevant since the errors between ordered classes do not hold the same importance. More metrics must be used during the learning and validation of the models. However, these other objectives may impair accuracy \cite{ben-david09}. Hence, a fair balance must be sought between the different needs of each problem.

Standard classifiers are discouraged for monotonic classification since they do not contemplate these constraints and their predictions violate the monotonicity required by certain applications. A classic example of these non-monotonic models is the standard decision tree \cite{ben-david95}. Standard $k$-Nearest Neighbors algorithm also does not take these restrictions into account, which may lead to further harm as a result of their presence in preprocessing techniques \cite{gonzalez19}.


In recent years, new algorithms have been designed to minimize the number of monotonic violations in their predictions \cite{ben-david95,gonzalez15,cano19}, i.e. the number of pairs of instances that break monotonicity \cite{ben-david95}. To do so, some approaches focus their entire learning mechanism just on monotonicity. This strategy usually achieves completely monotonic models, but it could lead to wrong generalizations being made that are different to the knowledge of the problem. Others infer monotonic relations from the training set while maximizing their accuracy. These models have been adapted from different families of classifiers \cite{cano19}, such as decision trees \cite{ben-david95,marsala15,pei18}, support vector machines \cite{chen14}, fuzzy model based classifiers \cite{alcala17,li15}, neural networks \cite{fernandez-navarro14,zhu17} and ensemble learning \cite{dembczynski09,qian15,gonzalez15}.


Instance-based learning has proven to be a good approach for monotonic classification \cite{ben-david92,duivesteijn08,lievens08,garcia16b}. However, some of these methods, such as Monotonic $k$-Nearest Neighbors \cite{duivesteijn08}  (M$k$NN), need to learn from a fully monotonic set to ensure monotonic predictions. This is rarely the case in real-life scenarios, where class noise and discrepancies are common. Therefore, data preprocessing \cite{garcia15,pan14,cano17,gonzalez19} and relabeling strategies \cite{potharst09,feelders10} must be used to remove non-monotonic samples or to change their class labels in order to force a monotonic set.



In standard classification, Fuzzy $k$-Nearest Neighbors \cite{keller85} is a very solid method with high performance, thanks to its high robustness to class noise \cite{derrac14}. This class noise robustness mainly lies in the extraction of the class memberships for the crisp training samples by nearest neighbor rule. In this process, the class memberships of noisy instances are shared with surrounding classes and the incorrectly assigned class looses its influence. However, these mechanisms do not consider monotonic constraints and Fuzzy $k$-NN cannot deal with monotonic violations or monotonic noise in the training set. 


In this paper, a new model designed on the basis of Fuzzy $k$-NN with notions of M$k$NN is proposed to take monotonic constraints into account, and is called Monotonic Fuzzy $k$-Nearest Neighbors (MonF$k$NN). MonF$k$NN has been designed with three desired features:
\begin{enumerate}[(i)]
    \item Robustness against monotonic violations.
    \item Monotonic predictions without a pure monotonic training set.
    \item Flexibility in its configurations covering different needs of performance.
\end{enumerate}

With these objectives in mind, MonF$k$NN has been designed with new mechanisms to manage monotonicity constraints and the monotonic violations in the training set. The main contributions of the MonF$k$NN design are:
\begin{enumerate}[(i)]
    \item The initial robustness of Fuzzy $k$-NN has been redesigned to mitigate the influence of monotonic violations. Firstly, the violations due to sample replicas with different classes are joined to form one class membership. Then, our approach incorporates a strictly monotonic nearest neighbor rule to the calculation of the memberships of the training examples. 
    \item These monotonically constrained memberships and their medians are used in the prediction phase. The class memberships aggregation of MonF$k$NN is also monotonically constrained by the nearest neighbor extraction or a penalty to the contribution of non-monotonic instances.  
    \item MonF$k$NN was built as a flexible classifier that covers different necessities of monotonicity and accuracy by tuning its parameters. It can be configured with a rigidly monotonic or standard $k$-NN rule if monotonicity or precision is preferred in the predictions, respectively.
\end{enumerate}


All these mechanisms reinforce the robustness of our proposal against monotonic noise without the need for relabeling. We understand monotonic noise as being the actual noise that can alter the class labels and, as a result, change the monotonic constraints among the samples in the data. Their parameters make our proposal adaptable to the different objectives of monotonic classification. We distinguish two different parameter configurations: a pure monotonic version in which monotonicity is prioritized, and an approximate configuration that focuses more on the prediction accuracy.



We have performed several empirical studies to verify the desired features of MonF$k$NN. First, different behaviors of its two configurations are empirically analyzed and compared to the original F$k$NN. Then, our proposal is compared with 7 methods from the state-of-the-art, exhibiting substantial improvements in accuracy and maintaining the best degree of monotonicity. Finally, the robustness of our method against monotonic noise, i.e. monotonic violations, is shown in contrast to M$k$NN. In this last experiment, MonF$k$NN performs considerably better than Monotonic $k$-NN in scenarios with large amounts of class noise. The experimental framework used consists of 12 data-sets commonly used in monotonic classification, 7 monotonic classifiers and 3 metrics covering different aspects of performance: Accuracy, Mean Absolute Error and Non-Monotonic Index. All results are additionally validated with the non-parametric statistical Wilcoxon and Friedman rank \cite{garcia08,garcia10} and Bayesian Sign tests \cite{benavoli17}.


The paper is organized as follows. In Section \ref{sec:background}, we present the problem of classification with monotonic constraints and the methods related to our proposal: M$k$NN and Fuzzy $k$-NN. Section \ref{sec:monfknn} is dedicated to explaining our model MonF$k$NN in detail and its algorithmic differences as compared to F$k$NN. The experimental framework used in the different empirical studies is presented in Section \ref{sec:framework}. In Section \ref{sec:experiments}, the previously mentioned empirical studies are carried out and analyzed. Finally, the main conclusions of this study are stated in Section \ref{sec:conclusions}.

\section{Preliminaries}
\label{sec:background}

In this section, we introduce the preliminaries needed: Classification with monotonicity constraints, Monotonic $k$-Nearest Neighbors and the original Fuzzy $k$-Nearest Neighbors. 

\subsection{Monotonic Classification}


Monotonic classification \cite{cano19} is an ordinal regression problem with monotonic constraints relating to the order of the variables and the class labels. Ordinal regression and/or classification can be seen as a nonstandard classification problem \cite{charte19}, which attempts to minimize the difference between the predicted labels and the real labels. Classification with monotonic constraints is also considered to be a nonstandard supervised learning problem \cite{charte19}.  

Formally, monotonic classification aims to predict the class label $y$ from input vector $x$ with $Q$ number of features, where $y \in \mathcal{Y}=\{l_1,l_2,\ldots,l_c\}$ and $x$ represents an individual of our classification problem. The categories $\mathcal{Y}$ are arranged in an order relation $\prec$ as $l_1\prec l_2\prec\ldots\prec l_c$. And, as the main property of monotonic classification, the attributes and class predictions are monotonically constrained by the problem prior-knowledge, i.e. $x\succeq x' \rightarrow f(x) \ge f(x')$ \cite{kotlowski13}, where $x \succeq x'$ implies $\forall_{j=1,\ldots,Q}, x_j \geq x'_j$, that is, $x$ dominates $x'$. Therefore, the main objective is to build classifiers that do not violate these constraints, otherwise known as monotonic classifiers.

Two different types of monotonic classifiers can be distinguished: approximate monotonic models, which minimizes the number of monotonic violations in their decisions and pure monotonic classifiers, whose predictions are always monotonic concerning the training and future examples. The latter is hard to achieve, particularly in real-life applications where the training data-sets are rarely purely monotonic. To be considered monotonic, all of the pairs of instances in a data-set must be monotonic \cite{ben-david92}: $x_{i}\succeq  x_{j}\:\rightarrow y_i\geq y_j, \forall_{i,j}$. 

\subsection{Monotonic \textit{k}-Nearest Neighbors}

M$k$NN \cite{duivesteijn08} modifies the standard nearest neighbor rule of the well-known lazy learning method to avoid monotonic violations in its predictions. To do so, M$k$NN computes for each new example $x_i$ the range $r_i = [y_{min}, y_{max}]$ of valid class labels, which satisfies the monotonic constraints. The lower-bound $y_{min}$ of $r_i$ is computed as the highest class label of all instances in the training set $\mathcal{D}$ below the example $x_i$. Analogously the upper-bound $y_{max}$ is the minimum class label of the instances in $\mathcal{D}$ that are higher than $x_i$ (see Eq. \ref{eq:knnRange}).

\begin{equation}
r_i =
\begin{cases}
&y_{min} = \max \{y \:|\:(x,y) \in \mathcal{D} \land x_i \succeq x \}\\
&y_{max} = \min \{y \:|\:(x,y) \in \mathcal{D} \land x \succeq x_i \}
\end{cases}
\label{eq:knnRange}
\end{equation}

Two different M$k$NN variants can be distinguished depending on how the neighbors are extracted for a new instance $x_i$. The \textit{InRange} variant considers the $k$ nearest examples $x_j$ with their class labels $y_j$ in the range $[y_{min}, y_{max}]$. The \textit{OutRange} version extracts first the $k$ nearest neighbors $x_j$ and then, those neighbors outside of the range $r_i$ are filtered out from the decision. If all of them are removed, a random label in $r_i$ is chosen. As in the standard $k$-NN method, the majority class among the $k$ neighbors is used as the predicted label.

M$k$NN is one of the methods that require monotonic data-sets to work properly \cite{duivesteijn08}. Since, with monotonicity violations, the range $r_i$ could not be correctly computed, a relabeling technique should be used to transform the non-monotonic training data into monotonic data. These techniques intend to identify and remove the monotonicity violations by making the fewest possible changes with minimum class difference. \cite{duivesteijn08,potharst09,feelders10}. 

\subsection{Fuzzy \textit{k}-Nearest Neighbors}

Fuzzy Sets \cite{zadeh96} express the uncertainty of the example memberships to each class label. The memberships of the example $x_i$ are represented as a degree of each class belonging $u_i =(u_{i1}, u_{i2}, \ldots,u_{ic})$, where $u_{il} \in [0,1]$ and $\displaystyle\sum^c_{l = 1} u_{il} = 1$. Nowadays, development in fuzzy sets and classifiers is still an ongoing process \cite{vskrjanc19}.

Fuzzy \textit{k}-Nearest Neighbors algorithms \cite{derrac14} incorporate fuzzy concepts into the classical $k$-NN decision rule to learn from fuzzy sets and produce fuzzy classification rules. Recently, different approaches have been proposed based on distinct fuzzy set extensions. However, the original Fuzzy $k$-NN \cite{keller85} (F$k$NN) is still one of the best approaches \cite{derrac14}. Recent approaches provide for the optimization of parameters in F$k$NN \cite{biswas18}.

For a given new instance $x_i$, Fuzzy $k$-NN \cite{keller85} extracts its $K$ nearest neighbors in the same manner as the standard $k$-NN. Then, its memberships for each class $l$ are computed with the following expression:

\begin{equation}
u(x,l) = \dfrac{  \displaystyle\sum_{j=1}^{K} u(x_j,l)*\dfrac{1} {||x-x_j||^{(m-1)}}  }{\displaystyle\sum_{j=1}^{K} \dfrac{1} {||x-x_j||^{(m-1)}}}\\
\label{eq:probAggr}
\end{equation}

As shown in Eq. \ref{eq:probAggr}, the membership $u(x_i,l)=u_{il}$ of sample $x_i$ to class $l$ is assigned with the product of the class membership $u(x_j,l)$ of the neighbors $x_j$ and the inverse of their distances to $x_i$. The latter serves as a weight that biases towards the memberships of nearer samples. The parameter $m$ determines the degree of influence of the neighbor distances. The recommended value $m=2$ \cite{keller85} makes the contributions of the neighboring samples reciprocal to their distances. A crisp class label for the example $x_i$ can be decided as being the label $l$ with the greatest membership degree $u_{il}$.

Facing a labeled training set, Fuzzy $k$-NN \cite{keller85} brings it into a fuzzy set with sample memberships using the nearest neighbor rule. For each training sample $x_i$, $k$ nearest neighbors are extracted using the leave-one-out scheme. Then memberships $u(x_i,l)$ for each class $l$ are computed according to Eq. \ref{eq:probFuzzied} with the number of neighbors $nn_l$ found for each class $l$. This transformation has proven useful against noisy samples as the memberships lose influence as they are spread to the surrounding classes (not the assigned class).

\begin{equation}
u(x_i,l) =
\begin{cases}
&0.51 + 0.49*(nn_l/k) \:, \quad \text{if} \quad  y_i = l \\
&0.49*(nn_l/k) \:, \quad \text{otherwise}
\end{cases}
\label{eq:probFuzzied}
\end{equation}

\section{Monotonic Fuzzy \textit{k}-Nearest Neighbors}
\label{sec:monfknn}

In this section, we explain our approach in detail -- MonF$k$NN and all its mechanisms that consider monotonicity constraints. In Subsection \ref{subs:medians}, we explain how MonF$k$NN gives a final class from class memberships in a more proper manner according to monotonicity. Subsection \ref{subs:classMem} is dedicated to the extraction of the class memberships from the training set and redesigned to reduce the impact of monotonic noise without the need for monotonic relabeling. In Subsection \ref{subs:memAggr}, the class membership aggregation built-in MonF$k$NN is explained and related to the robustness and flexibility of the classifier using its parameters. Finally, we discuss the algorithmic differences between our proposal and the original F$k$NN in Subsection \ref{subs:MFKNNvsFKNN}.


\subsection{From class memberships to the final class label}
\label{subs:medians}

Since F$k$NN works with class memberships, a mechanism that respects monotonicity is needed to get a final class from a vector whose elements sum up to the value of one. The class with the greatest membership is the most common decision in multiple classifiers. The original Fuzzy $k$-NN gives their crisp predictions as the class label with the highest membership. 

However, this might not be appropriate for scenarios with monotonic constraints. For example, let $x_i \leq x_j$ and their class memberships $u_i=(0.2, 0.2, 0.4, 0.2, 0.0)$ and $u_j=(0.0, 0.4, 0.3, 0.2, 0.1)$, then their final classes chosen with the highest membership break the monotonicity: $argmax(u_i) = l_3 > l_2 = argmax(u_j)$. Even though, the instance $x_j$ has more weight values assigned to the higher labels than instance $x_i$. In fact, $u_j$ weakly dominates $u_i$ according to the \textit{first degree stochastic dominance relation} (FSD) \cite{levy15} since the $x_i$ cumulative distribution function $U_i=(0.2, 0.4, 0.8, 1.0, 1.0)$ is greater, element by element, than $U_j=(0.0, 0.4, 0.7, 0.9, 1.0)$, that is, $u_i \preceq_{FSD} u_j  \Longleftrightarrow (\forall l \in \mathcal{Y})(U_i(l) \geq U_j(l))$. To make FSD applicable, class membership vectors are normalized to sum up to the value of one and treated as probability mass functions. Therefore, a cumulative distribution function U can be computed for given normalized class memberships, where FSD is defined. This transformation can be done thanks to the order relation between classes in monotonic classification. FSD is useful for defining monotonicity constraints in probabilistic classifications \cite{lievens08,lievens10}, with the expression $x_i \leq x_j \implies u_i \preceq_{FSD} u_j$.

Therefore, the function that transfers a membership vector to a class label must satisfy $u_i \preceq_{FSD} u_j\implies y_i \leq y_j$. Centrality measures, such as mean and median, have proven to be good solutions \cite{levy15,lievens08}. Particularly, the median is applicable to ordinal problems. Following the traditional definition of median as the 50th percentile, the median is computed as the range [$l_m$, $l_M$]:

\begin{equation}
\begin{aligned}
&l_m = \min \{l \in \mathcal{Y} \:|\: U\{ X \leq l \} \geq 1/2 \}\\
&l_M = \max \{l \in \mathcal{Y} \:|\: U\{ X \geq l \} \geq 1/2 \}
\end{aligned}
\label{eq:medianRange}
\end{equation}

\noindent where $l$ is a class label of possible labels $\mathcal{Y}$, $U\{ X \leq l \}$ is the cumulative membership/probability of belonging to a class smaller or equal to $l$ and $U\{X \geq l \}$ is the analogous definition for a class greater or equal to $l$.

Going back to the previous example, the classes for $x_i$ and $x_j$ chosen by the median does not break monotonicity: $med(u_i) = med(u_j) = 3$. For $l_m \neq l_M$, any class label $l$ which $l_m < l < l_M$ must have a membership $u(l) = 0$ and $U(l_m) = U(l_M) = 1/2$. For example, instance $x_t$ with class memberships $u_t=(0.2, 0.3, 0, 0.3, 0.2)$ could be assigned to the classes $med(u_t)= [2, 4] = 3$.

\subsection{Class memberships robust to monotonic noise}
\label{subs:classMem}

In this subsection, the class membership calculation redesigned to monotonic classification is explained. The objective of this first stage is to fix or reduce the influence of non-monotonic examples in the classification. Our method uses the robustness of the traditional Fuzzy $k$-NN within the knowledge of the monotonic relations between the neighbors. Algorithm \ref{alg:classMemberships} summarizes the procedure of obtaining robust noise class memberships for the training set.

\begin{algorithm}
\caption{Training class memberships extraction}
\label{alg:classMemberships}

\begin{algorithmic}[1]
\Function{TrainClassMemberships}{$\{\mathcal{D},y\}$ - Training data-set, k - Nearest neighbors considered,  $\RCr$ - Real Class relevance}
	\For{$x_i \in \mathcal{D} $} \label{ln:startDuplicated}
	    \For{$l \in \mathcal{Y}$}
	        \If{$x_i$ duplicated-in $\mathcal{D}$}
	            \State Compute $u(x_i,l)$ with \textbf{expression \ref{eq:probDoubts}}
            \Else
            \State $u(x_i,l) =
                \begin{cases}
                &1 \quad  y_i = l \\
                &0
                \end{cases}$
            \EndIf
        \EndFor
    \EndFor
    
    \State $\mathcal{D}' = removeDuplicates(\mathcal{D})$
    \For{$x_i \in \mathcal{D}' $}
        \State $y_i' = med(u_i)$ \label{ln:endDuplicated}
        \Comment{See \textbf{expression \ref{eq:medianRange}} $\quad \quad \quad \quad$}
    \EndFor
    \For{$x_i \in \mathcal{D}' $}
        \If{$x_i$ not-duplicated-in $\mathcal{D}$}
        \label{ln:monKNN}
        \State Compute range $r_i$ with $(\mathcal{D}',y')$ and \textbf{expression \ref{eq:knnRange}} \label{ln:rangeMem}
        \State \Comment{See \textbf{Algorithm \ref{alg:neighborsAsMonKNN}} $\quad \quad \quad \quad$}
        \State $nn =$ neighborsAsM$k$NN$(x_i,r_i,k,inRange,\mathcal{D}',y')$
            \For{$l \in \mathcal{Y}$} \label{ln:MemFusion}
                \State Compute $u(x_i,l)$ with \textbf{expression \ref{eq:probMFuzzied}}
                
        
            \EndFor
        \EndIf
    \EndFor \label{ln:endMemberships}
    \State \textbf{output:} $(\mathcal{D}',u)$
 \EndFunction

 \end{algorithmic}
 \end{algorithm}

First, we have to deal with the simplest monotonic violations, that is, instances with the same input values and different classes (Lines \ref{ln:startDuplicated}-\ref{ln:endDuplicated} of Algorithm \ref{alg:classMemberships}). These mislabels frequently appear in traditional data-sets \cite{ben-david92} of classification with monotonic constraints as these sets are rankings or evaluations made by different experts. 

Therefore, MonF$k$NN first substitutes the replicas of any example $x$ with one feature vector $x$ and its memberships $u(x)$. The membership $u(x,l)$ of the instance $x$ to the class $l$ is computed with the frequency of duplicated examples $x_j$ in the training set $\mathcal{D}$ belonging to class $l$ ($y_j = l$), as shown in the following expression:

\begin{equation}
u(x,l) = \dfrac{|\{x_j \in \mathcal{D} | x_j = x \land y_j = l\}|} {|\{x_j \in \mathcal{D} | x_j = x\}|}\\
\label{eq:probDoubts}
\end{equation}

The class label of an instance $x$ after the elimination of its replicas is obtained by the median of the resulting memberships, as shown in Line \ref{ln:endDuplicated} of Algorithm \ref{alg:classMemberships}. However, this vector will be used in the classification function with the membership aggregation as stated in the next subsection.

Then, MonF$k$NN estimates the memberships of the remaining instances, which corresponds to Lines \ref{ln:endDuplicated} - \ref{ln:endMemberships} of Algorithm \ref{alg:classMemberships}. This estimation is made using the information of the nearest neighbors of each instance. However, these nearest neighbors are extracted with a monotonic nearest neighbor rule (M$k$NN) instead of a traditional rule as we aim for memberships that respect monotonic constraints as much as possible. Algorithm \ref{alg:neighborsAsMonKNN} exemplifies the extraction of these monotonically constrained neighbors for a given instance $x$ as in M$k$NN.

\begin{algorithm}
\caption{Monotonic nearest neighbor rule}
\label{alg:neighborsAsMonKNN}

\begin{algorithmic}[1]
\Function{NeighborsAsM$k$NN}{$x$ - tested sample, $r$ - range of valid classes, $k$ - considered neighbors, $typeRange$ - inRange or outRange, $\{\mathcal{D},y\}$ - Training data-set}
    \State \textbf{initialize:} $nn=\{\}$ 
	\For{$x_i \in \mathcal{D}$}
	    \If{$typeRange == outRange$ \textbf{or} $y_i \in r$}
	        \If{$Size(nn) < k$}
	           \State Insert $x_i$ in $nn$
	        \Else
	            \State $x_{max} = \arg \max_{x_j \in nn} ||x-x_j||$
	            \If{$||x-x_i|| < ||x-x_{max}||$}
	                \State Replace $x_{max}$ by $x_i$ in $nn$
	            \EndIf
	        \EndIf
	    \EndIf
	\EndFor
	
	\State \textbf{output:} $nn$
\EndFunction

\end{algorithmic}
\end{algorithm}

In this case, Algorithm \ref{alg:neighborsAsMonKNN} is configured as an \textit{inRange} variant as pointed out in Line \ref{ln:monKNN} of Algorithm \ref{alg:classMemberships}. That is, the nearest neighbors of an example $x_i$ are constrained to a range $r_i = [y_{min}, y_{max}]$ of possible classes (Line \ref{ln:rangeMem}), which preserves the monotonicity of the data-set.





Once the nearest neighbors for each example $x_i$ are obtained, the information of the neighbor classes is fused into $x_i$ class memberships (Line \ref{ln:MemFusion}). For an instance $x_i$, the membership $u(x_i,l)$ to class $l$ is computed with the following expression: 

\begin{equation}
u(x_i,l) =
\begin{cases}
&\RCr + (nn_l/k)*(1-\RCr) \quad \text{if} \quad  y_i = l \\
&(nn_l/k)*(1-\RCr)
\end{cases}
\label{eq:probMFuzzied}
\end{equation}

\noindent where $nn_l$ is the number of nearest neighbors of the class $l$, $k$ the total number of neighbors extracted for instance $x_i$ and $y_i$ is the original class label of the example $x_i$. $\RCr$ is a new parameter called \textit{"Real Class relevance"}.

Apart from the use of the monotonic nearest neighbor rule, the inclusion of $\RCr$ is another main difference between our approach MonF$k$NN and the original Fuzzy $k$-NN. $\RCr$ can be seen as the minimum membership assigned to original class $y_i$ of the instance $x_i$, in case there are no neighbors labeled with $y_i$. In F$k$NN, $\RCr$ corresponds to the value of $0.51$, that is, every instance maintains its real class, even those noisy examples surrounded by other classes. By being a parameter, our method lets the user control the treatment of monotonic noise.


There are some values for $\RCr$ in the range $[0,1]$ that have very interesting and distinct behaviors. In the case of a really noisy data-set where no labels can be trusted, $\RCr$ could be set to 0. This leaves all the responsibility to the calculation of the range of valid classes $r_i$ and the nearest neighbors. In the presence of instances with the same input values and different classes, the user could choose only to treat them with $\RCr=1$. Finally, if practitioners want to consider the originally labeled instances, we recommend assigning $\RCr$ to $0.5$. This value ensures that the actual class is within the set of medians. In contrast to Fuzzy $k$-NN and its $0.51$, if all neighbors belong to a same single class that is different to the current class, our method forces to choose in between these two classes. Usually, this last value ($\RCr=0.5$) is a good trade-off, mainly stable and with better performance.

During this process, the impact of monotonic inconsistencies will be either reduced or fixed. The inconsistencies of instances with the same input vectors and different classes are completely fixed by being substituted by only a sample and class memberships with the information of their different classes. The mislabeled samples, i.e. noisy or non-monotonic examples, will have less influence towards their noisy class as they will be surrounded by more appropriated classes and their class memberships will be shared into classes in which they fit monotonically. This is the first mechanism of our method to alleviate the presence of monotonic violations, without the need for relabeling.


\subsection{Flexible membership aggregation}
\label{subs:memAggr}

After estimating the class memberships of every training instance, our algorithm is ready to predict new examples. This last phase has been designed to cover different needs of monotonic scenarios. In addition to the control of noise treatment, greater flexibility has been sought, allowing users to choose between more accurate or pure monotonic predictions.

Algorithm \ref{alg:MonFkNN} represents in pseudo-code the whole prediction procedure of our proposal MonF$k$NN. Particularly, the prediction of a new instance $x_i$ is detailed after having previously computed the monotonically-constrained class memberships of the training set as the previous Algorithm \ref{alg:classMemberships} is referred in Line \ref{ln:trainClassMem}.

\begin{algorithm}[H]
\caption{MonF$k$NN: Prediction stage}
\label{alg:MonFkNN}

\begin{algorithmic}[1]
\Function{MonF$k$NN}{$x_i$ - sample to predict, $\{\mathcal{D},y\}$ - training data-set, $k$ - neighbors considered for training class memberships,  $\RCr$ - Real Class relevance, $K$ - neighbors considered for prediction, $typeRange$ - inRange or outRange, $\pOR'$ - out-of-range penalty}
    
	\State $(\mathcal{D}',u') = $ TrainClassMemberships$(\mathcal{D}, y, k, \RCr)$ \label{ln:trainClassMem}
	\State Obtain medians $y'$ of each sample in $\mathcal{D}'$ with $u'$ and \textbf{expression \ref{eq:medianRange}} \label{ln:mediansMonFkNN} 
	
	\State Compute range $r_i$ with \textbf{expression \ref{eq:knnRange}} and $(\mathcal{D}',y')$ \label{ln:rangeMonFkNN}
	\State \Comment{See \textbf{Algorithm \ref{alg:neighborsAsMonKNN}} $\quad \quad \quad \quad$}
	\State $nn =$ neighborsAsM$k$NN$(x_i,r_i,K,typeRange,\mathcal{D}',y')$ \label{ln:asMonkNN}
    \For{$x_j \in nn$} \label{ln:applyPOR1}
        \If{$typeRange == inRange$ \textbf{or} $y_j \in r_x$}
            \State $\pOR_j = 1$
        \Else  \Comment{Neighbors $nn$ out of range $r_x$ are penalized with $\pOR'$}
            \State $\pOR_j = \pOR'$
        \EndIf
    \EndFor \label{ln:applyPOR2}
    
    \State Compute class memberships $u_i$ of $x_i$ with \textbf{expression \ref{eq:probMAggr}} 
    
    \State \textbf{output:} $med(u_i)$
 \EndFunction

 \end{algorithmic}
 \end{algorithm}

As shown in Line \ref{ln:asMonkNN}, MonF$k$NN embeds another M$k$NN (Algorithm \ref{alg:neighborsAsMonKNN}) to obtain the neighbors used in the membership aggregation and final prediction. This M$k$NN also has two versions, \textit{inRange} and \textit{outRange} versions. They are, however, substantially different when compared to original variants.

The \textit{inRange} alternative is based on the same idea of the original M$k$NN, where the neighbors of an example must belong to a set of monotonically valid classes. However, this range of classes is obtained using the medians acquired from the class memberships of the training instances constrained by monotonicity, as seen in Line \ref{ln:mediansMonFkNN} and Line \ref{ln:rangeMonFkNN}. This breakthrough improves our method by increasing monotonic noise robustness. Firstly, an \textit{inRange} nearest neighbor rule removes monotonic inconsistencies in the known data-set as previously shown in Algorithm \ref{alg:classMemberships}. Then, the second M$k$NN uses this fixed training set $(\mathcal{D}',y')$ to give monotonic predictions as seen in Algorithm \ref{alg:MonFkNN}. 

The \textit{outRange} version of our method is completely different from the previous \textit{outRange} rule. It has been designed with the intention of prioritizing to some extent the predictive ability of the classifier over monotonicity. With this purpose in mind, our method considers any example as a valid neighbor regardless of its class label. In contrast to the original model, no filtering or removal of neighbors outside the valid range is performed. However, their relevance in the membership aggregation can be reduced if needed, thanks to a penalty factor introduced in the aggregation expression.

Then, for a new example $x$, its nearest neighbors are obtained according to the chosen variant. Their memberships are aggregated with the original F$k$NN formula with the addition of the penalty factor for the \textit{outRange} version. The following expression shows how this parameter is integrated:

\begin{equation}
u(x,l) = \dfrac{  \displaystyle\sum_{j=1}^{K} u(x_j,l)*\dfrac{\pOR_j} {||x-x_j||^{(m-1)}}  }{\displaystyle\sum_{j=1}^{K} \dfrac{\pOR_j} {||x-x_j||^{(m-1)}}}\\
\label{eq:probMAggr}
\end{equation}

As previously, the membership $u(x,l)$ of the new sample $x$ to the class label $l$ is the result of the sum of the class memberships $u(x_j,l)$ of the neighbors $x_j$ inversely weighted with their distance to $x$. In the \textit{outRange} version of our method, there is another weighting factor in the contribution to the final memberships, the parameter referred to as \textit{"penalty of outRange"} ($\pOR$). The factor $\pOR_j$ is applicable only if the class $y_j$ of the neighbor $x_j$ is not in the valid class range $r_x$ of $x$ as exemplified in Lines \ref{ln:applyPOR1} to \ref{ln:applyPOR2} . It can be configured with continuous values from 0 to 1. When it is assigned to 1, no penalty is applied. The value 0 means a full penalty, that is, neighbors with invalid classes will not participate in the membership aggregation. For all practical purposes, this last behavior is equivalent to the \textit{outRange} M$k$NN. We recommend using 0.5 since it is a good balance between reducing their relevance and considering them in the decision. 

Finally, the class prediction of the new example $x$ is the median of the resulting normalized class memberships. 

As presented, MonF$k$NN has been developed to be robust to monotonic noise and versatile in many scenarios. The two versions \textit{inRange} and \textit{outRange} with the parameter $\pOR$ and the previously mentioned $\RCr$ help to tune the algorithm according to the necessities of different kinds of problems. 

Among the possibilities that offer these parameters, we have named two configurations with very distinctive behaviors: Pure Monotonic (MonF$k$NN-PM or PM) and Approximate Monotonic (MonF$k$NN-AM or AM) Fuzzy $k$-NN. The Pure Monotonic configuration corresponds to a value of 0.5 for the $\RCr$ parameter and the use of \textit{inRange} rule to obtain the memberships of new instances. This approach aims to give predictions with the minimum violations of monotonicity. In every part of the algorithm, it prioritizes monotonicity over very accurate predictions.

MonF$k$NN-AM prioritizes the predictive ability and relaxes the monotonic constraints. The memberships of the training set are obtained by the treatment of samples with the same feature values and different classes. Those unique examples will have a membership of 1 to the actual class and 0 for the rest. This behavior is achieved with $\RCr = 1$. Then, as we are looking for more accurate predictions, all instances can be considered to be valid neighbors and to contribute to the final aggregation. Those instances with invalid class labels, however, will contribute with only half of their class memberships ($\pOR = 0.5$).

Our proposal MonF$k$NN is available at the GitHub Repository\footnote{\url{https://github.com/sergiogvz/MonFkNN}}.

\subsection{Differences between standard F$k$NN and MonF$k$NN: Theoretical discussion}
\label{subs:MFKNNvsFKNN}

Standard F$k$NN and MonF$k$NN have a similar mathematical formulation. In other words, the expressions used by MonF$k$NN in the training class membership extraction (Eq. \ref{eq:probMFuzzied}) and in the membership aggregation (Eq. \ref{eq:probMAggr}) are the same as those used by F$k$NN (Eq. \ref{eq:probFuzzied} and Eq. \ref{eq:probAggr}), for $\RCr=0.51$ and $\pOR=1$. The global behavior of our method is however still completely different to the standard F$k$NN, due to significant algorithmic differences. Table \ref{tab:MFKNNvsFKNN} summarizes the main differences between standard F$k$NN and our proposal MonF$k$NN.

\begin{table}[ht]
    \centering
    	\resizebox{\textwidth}{!}{
    \begin{tabular}{cc}
    \toprule
         \textbf{F$k$NN} & \textbf{MonF$k$NN}\\
         \midrule
        No special treatment of duplicates. & Duplicates are reduced to a single instance.\\
        Standard nearest neighbor rules. & Monotonic nearest neighbor rules.\\
        Standard training membership extraction. & Monotonically constrained class memberships.\\
        Conservation of original classes in the & Loss of influence of original class towards  \\
        training class membership extraction.  &  monotonicity with $\RCr<=0.5$.\\
        Value 0.51 in Eq. \ref{eq:probFuzzied}   & Parameter $\RCr$ in Eq. \ref{eq:probMFuzzied}  \\
        Standard class membership aggregation. & Monotonically constrained membership aggregation.\\
        No penalty to any neighbors in Eq. \ref{eq:probAggr} & $\pOR$ Penalty to out-of-range neighbors in Eq. \ref{eq:probMAggr}.\\
        Final class as highest membership &  Final class as median of class memberships\\
         \bottomrule
    \end{tabular}
    }
    \caption{Summary of algorithmic differences between standard F$k$NN and MonF$k$NN.}
    \label{tab:MFKNNvsFKNN}
\end{table}

Each of the differences mentioned in Table \ref{tab:MFKNNvsFKNN} is described and explained below:
\begin{itemize}
    \item The data-set used to compute the training class memberships is modified before applying the neighborhood rule. The inconsistencies of duplicates are eliminated and reduced to a single instance. The classes of the resultant instances are assigned to the median calculated with the frequency of the appearance of duplicates for each class. This procedure could not even be considered in standard classification, where there is no ordering relationship between classes.
    \item The neighborhood considered for each training instance is constrained to the monotonicity of the data-set. Then, their resultant class memberships are also monotonically constrained. These adaptations completely modify the neighbors contributing in Eq. \ref{eq:probFuzzied} and the whole procedure. In addition, the value of $0.51$ for $\RCr$ is discouraged in MonF$k$NN in favor of $0.5$ due to its contribution to the medians of the samples, above-mentioned in Section \ref{subs:classMem}.
    \item The original F$k$NN and MonF$k$NN also share the same membership aggregation, i.e. their expressions (Eq. \ref{eq:probFuzzied} and Eq. \ref{eq:probMFuzzied}) are the same for \textit{InRange} and \textit{outRange} (with $\pOR=1$) versions of MonF$k$NN. However, their behavior and their predictions are completely different, due to the differences in the nearest neighbor rule, in the training set and class memberships used in the aggregation procedure. As previously explained, the training class memberships extraction of MonF$k$NN modifies the training set fixing some monotonic inconsistencies. Duplicates are removed and some training samples might change their classes to preserve the monotonicity of the data-set.
    \item In MonF$k$NN, the classes of the training samples determine the monotonically valid classes of the unlabeled instances. Thus, training samples with classes not valid for an instance $x$ will be discarded from the neighborhood (\textit{inRange} version) or penalized with the parameter $\pOR$ (\textit{outRange} version). The configuration \textit{outRange} version with $\pOR = 1$ is also discouraged since the final purpose of MonF$k$NN is to take monotonic constraints into consideration, at least to some extent.
    \item These mechanics acquire different neighbors to those drawn by F$k$NN for the same test sample, that is, different class memberships and prediction. Finally, the median as the final class of the class membership vector already implies a significant change in the behavior of the method. 
\end{itemize}

These differences between our proposal and the traditional F$k$NN are clearly supported by the experiments carried out in Section \ref{subs:exp-FKNN}.

\section{Experimental framework}
\label{sec:framework}

This section is devoted to introducing the experimental framework used in the different empirical studies of the paper. In our experiments, we have included 12 data-sets of a good variety of problems presenting real monotonic constraints. The data-sets can be seen in Table \ref{tab:data-sets}, where the number of instances, attributes and classes are detailed for each data-set in the column Ins., At. and Cl., respectively. The column At. Directions indicates the monotonic direction of the relationship between each attribute and the class: direct (+) or inverse monotony (-). This information is extracted from the description of the problems involving the data-sets. The column Comparable Pairs shows the percentage of pairs of comparable samples over the total number of pairs. Two instances $x_i$ and $x_j$ are comparable if their inputs have an order relation, i.e. $x_i \succeq  x_j$ or $x_i \preceq x_j$. On average, one-third of the total number of pairs of these data-sets are comparable and potential violations of monotonicity in the classification process. This quite large amount cannot be neglected.

These data-sets are chosen as the most frequently used in the monotonic classification literature. The classical monotonic set \textit{ERA}, \textit{ESL}, \textit{LEV} and \textit{SWD} \cite{ben-david92} are also considered in the study. Additionally, the data-set \textit{artiset} is employed for a comparative study on monotonic noise robustness of MonF$k$NN (see Subsection \ref{subs:exp-noise}). \textit{Artiset} is an artificial data-set with two attributes ($x_1, x_2$) and $nCl$ number of classes. For attributes $x_1, x_2 \in [0,1]$, the class is computed as the truncation of the outcome of the following formula:

\begin{equation*}
 f(x_1,x_2) = (x_1 + \frac{x_2^2 - x_1^2}{2})*nCl
\end{equation*}


A 10-fold cross-validation scheme (10-fcv) is carried out to run the different classifiers over these sets. Their partitions have been extracted from the KEEL repository \cite{keel17}.

\begin{table}[!htp]
	\centering
	\caption{Description of the 12 data-sets used.}
	\label{tab:data-sets}
	\resizebox{\textwidth}{!}{
	\begin{tabular}{cccccc}
	\toprule  
    Data-set & Ins.  & At. & Cl. & At. Directions & Comparable Pairs\\
	\midrule  
	\textit{artiset} & 1000 & 2 & 10 & All direct directions & 49.79\% \\
	\textit{balance} & 625  & 4   & 3 & \{-, -, +, +\} & 25.64\%\\
	\textit{bostonhousing4cl} & 506 & 13 & 4 & \{-, +, -, +, -, +, -, +, -, -, -, +, -\} & 14.85\%\\
    \textit{car} & 1728   &6  & 4  & All direct directions & 14.36\%\\
	\textit{ERA} &  1000 & 4   &  9 & All direct directions & 16.77\%\\
	\textit{ESL}  & 488  & 4  &  9 & All direct directions & 70.65\%\\
	\textit{LEV}  & 1000  & 4 & 5  & All direct directions & 24.08\%\\
    \textit{machineCPU} & 209 & 6 & 4 & \{-, +, +, +, +, +\} &  49.53\%\\
    \textit{qualitative\_bankruptcy} & 250 & 6 & 2 & All inverse directions & 43.77\%\\
	\textit{SWD} & 1000  & 10  &  4 & All direct directions & 12.62\%\\
    \textit{windsorhousing} & 546 & 11 & 2 & All direct directions &   27.07\%\\
    \textit{wisconsin}  & 683 & 9 & 2 & All direct directions & 58.04\%\\
	\bottomrule
	\end{tabular}
	}
\end{table}

The classifiers involved in the empirical comparisons are:
\begin{itemize}
    \item Monotonic $k$-NN (M$k$NN) \cite{duivesteijn08}
    \item Ordinal Stochastic Dominance Learning (OSDL) \cite{lievens08}
    \item Ordinal Learning Module (OLM) \cite{ben-david92}
    \item Monotonic Multi-Layer Perceptron network (MonMLP) \cite{lang05}
    \item C4.5 decision tree for monotonic induction (MID) \cite{ben-david95}
    \item Rank Discrimination Measure Tree (RDMT) \cite{marsala15}
    \item Partially Monotonic Decision Tree (PMDT) \cite{pei18}
\end{itemize}

Table \ref{tab:parameters} details the parameters chosen according to the recommendations found in the original papers. As a requirement of M$k$NN, a relabeling technique \cite{feelders10} is applied to training data-sets before fitting M$k$NN. On the contrary, the rest of the algorithms, including MonF$k$NN, do not need this relabeling procedure. Therefore, all the results shown for M$k$NN are obtained with relabeled training sets, while other methods are trained with the original training data-sets.

\begin{table}[ht]
	\centering
	\caption{Parameters considered for the algorithms compared.}
    \label{tab:parameters}
	\resizebox{\textwidth}{!}{
		\begin{tabular}{ll}
			\toprule
			Algorithm & Parameters\\
            \midrule
			M$k$NN \cite{duivesteijn08} & $k$ = 5, distance = euclidean, neighborsType = inRange\\
            OSDL \cite{lievens08} & balanced = No, classificationType = median,\\
			&lowerBound = 0, upperBound = 1\\
			&tuneInterpolationParameter = No, weighted = No,  \\
			&interpolationStepSize = 10, interpolationParameter = 0.5\\
			OLM \cite{ben-david92} & modeResolution = conservative\\
            & modeClassification = conservative\\
            MonMLP \cite{lang05} & default parameters, hidden1 = 8  \\
            & iter.max = 1000, monotonic = all att\\
			MID \cite{ben-david95} & R = 1, confidence = 0.25, items per leaf = 2 \\
			RDMT \cite{marsala15} & H = Pessimistic rank discrimination measure, \\
			& measureThreshold = 0, items per leaf = 2\\
			PMDT \cite{pei18} & threshold $\theta$ = 0, items per leaf = 2\\
			F$k$NN \cite{keller85} & $k$ = 5, $K$ = 9,  distance = euclidean\\
            MonF$k$NN & $k$ = 5, $K$ = 9,  distance = euclidean\\
            \hspace{1pt} Pure Monotonic & $\RCr$ = 0.5, neighborsType = inRange\\
            \hspace{1pt} Approximate Monotonic & $\RCr = 1$, neighborsType = outRange, $\pOR = 0.5$\\
            \bottomrule
		\end{tabular}
	}
\end{table}

In order to evaluate the classifiers' proficiency, we have employed three measures of different aspects of their performance: predictive capability, error cost and monotonicity. Standard accuracy is used to evaluate the predictive capability of the models. Mean Absolute Error (MAE) is computed as the average differences of the true instance ranks and the predicted ranks. To evaluate monotonicity, Non-Monotonic Index (NMI) \cite{cano19} measures the ratio of pairs of samples (NMP) that break monotonicity among the total of pairs, with $N$ being the number of samples in the data-set:
\begin{equation*}
 \NMI = \frac{\NMP}{N^2-N}
\end{equation*}



These measures are computed over a set merged from the test predictions of 10-fcv sets for each data-set and classifier. Finally, the Wilcoxon statistical test, Friedman rank test \cite{garcia08,garcia10} with Holm post-hoc procedure \cite{holm79} and Bayesian Sign test \cite{benavoli17} are used to validate the results of the empirical comparisons. In the Bayesian Sign test, a distribution of the differences of the results achieved by methods $A$ and $B$ is computed thanks to the Dirichlet Process. This distribution is shown in a graphical space divided into 3 regions: left, rope and right. The location of the majority of distribution in these sectors indicates the final decision of the pairwise Bayesian non-parametric sign test: superiority of algorithm $B$ (left sector), statistical equivalence (rope sector) and superiority of algorithm $A$ (right sector). For the accuracy and MAE results, we have set the inferior and superior limit of the rope region to $-0.01$ and $0.01$, respectively.  However, we have adjusted the limits to $-0.0001$ and $0.0001$ for NMI since NMI values tend to be significantly smaller due to the big difference between the numbers of comparable instance pairs and all possible pairs. The R package rNPBST \cite{carrasco17} has been used to extract the graphical representations of the Bayesian Sign tests analyzed in the following empirical studies.

\section{Results and analysis}
\label{sec:experiments}

This section presents the results of the empirical studies and their analyses. First, the two configurations of MonF$k$NN are compared in Subsection \ref{subs:exp-FKNN}, showing their different strengths. Then, our proposal is compared to methods from the state-of-the-art in terms of prediction capability and monotonicity in Subsection \ref{subs:exp-State-mon} and Subsection \ref{subs:exp-State-mon}, respectively. In Subsection \ref{subs:exp-noise}, the last experiment tests the noise robustness of MonF$k$NN in contrast to M$k$NN.

\subsection{Evaluation of Monotonic Fuzzy k-NN approaches. Pure Monotonic vs Approximate Monotonic}
\label{subs:exp-FKNN}

A comparison between the Pure and Approximate Monotonic version of MonF$k$NN stresses the different behaviors and aspects of their performance. Additionally, the performance differences between the original F$k$NN and MonF$k$NN are analyzed. Table \ref{tab:fknns} shows the results of F$k$NN and the two configurations of our proposal MonF$k$NN in terms of Accuracy, MAE and NMI. Bold-face font indicates the best results obtained for each data-set and metric. 

\begin{table}[hbt]
  \centering
  \caption{Results for the Pure and Approximate Monotonic Fuzzy $k$-NN}
  \resizebox{\textwidth}{!}{
    \begin{tabular}{lrrr|rrr|rrr}
    \toprule
    & \multicolumn{3}{c}{\textbf{Accuracy}} & \multicolumn{3}{c}{\textbf{MAE}} & \multicolumn{3}{c}{\textbf{NMI}}\\
    \midrule
\textbf{}                        & \textbf{F$k$NN}   & \textbf{MonF$k$NN-PM} & \textbf{MonF$k$NN-AM} & \textbf{F$k$NN}   & \textbf{MonF$k$NN-PM} & \textbf{MonF$k$NN-AM} & \textbf{F$k$NN}   & \textbf{MonF$k$NN-PM} & \textbf{MonF$k$NN-AM} \\
\midrule
\textit{artiset}                 & 0.9339          & 0.9309              & \textbf{0.9349}     & 0.0661          & 0.0691              & \textbf{0.0651}     & \textbf{0.0000} & \textbf{0.0000}     & \textbf{0.0000}     \\
\textit{balance}                 & 0.8896          & \textbf{0.9307}     & 0.9008              & 0.1424          & \textbf{0.0853}     & 0.1168              & \textbf{0.0000} & \textbf{0.0000}     & 0.0001              \\
\textit{bostonhousing4cl}        & \textbf{0.7174} & 0.6561              & 0.7134              & \textbf{0.3241} & 0.3972              & 0.3261              & 0.0004          & \textbf{0.0000}     & 0.0001              \\
\textit{car}                     & 0.9311          & 0.9740              & \textbf{0.9834}     & 0.0793          & 0.0295              & \textbf{0.0195}     & 0.0002          & \textbf{0.0000}     & \textbf{0.0000}     \\
\textit{ERA}                     & 0.1730          & 0.2420              & \textbf{0.2430}     & 1.6660          & \textbf{1.2813}     & 1.2993              & 0.0141          & \textbf{0.0052}     & \textbf{0.0052}     \\
\textit{ESL}                     & 0.6783          & 0.7036              & \textbf{0.7131}     & 0.3484          & 0.3149              & \textbf{0.3053}     & 0.0014          & 0.0004              & \textbf{0.0003}     \\
\textit{LEV}                     & 0.6020          & \textbf{0.6377}     & 0.6110              & 0.4330          & \textbf{0.3927}     & 0.4223              & 0.0021          & \textbf{0.0004}     & 0.0009              \\
\textit{machineCPU}              & 0.6699          & \textbf{0.7033}     & 0.6699              & 0.3589          & \textbf{0.3158}     & 0.3493              & 0.0058          & \textbf{0.0002}     & 0.0017              \\
\textit{qualitative\_bankruptcy} & \textbf{0.9960} & \textbf{0.9960}     & \textbf{0.9960}     & \textbf{0.0040} & \textbf{0.0040}     & \textbf{0.0040}     & \textbf{0.0000} & \textbf{0.0000}     & \textbf{0.0000}     \\
\textit{SWD}                     & 0.5350          & 0.5807              & \textbf{0.5833}     & 0.5180          & \textbf{0.4370}     & 0.4380              & 0.0027          & 0.0007              & \textbf{0.0003}     \\
\textit{windsorhousing}          & \textbf{0.7857} & 0.7576              & 0.7839              & \textbf{0.2143} & 0.2424              & 0.2161              & 0.0062          & \textbf{0.0005}     & 0.0051              \\
\textit{wisconsin}               & \textbf{0.9678} & 0.9653              & 0.9663              & \textbf{0.0322} & 0.0347              & 0.0337              & \textbf{0.0000} & \textbf{0.0000}     & \textbf{0.0000}     \\
\midrule
\textit{Avg:}                    & 0.7400          & 0.7565              & \textbf{0.7583}     & 0.3489          & 0.3003              & \textbf{0.2996}     & 0.0027          & \textbf{0.0006}     & 0.0012 \\            
\bottomrule
    \end{tabular}%
    }
  \label{tab:fknns}%
\end{table}%

In Table \ref{tab:fknns}, the differences between both approaches (PM and AM) can be seen clearly. Just as they were designed, MonF$k$NN-AM has better accuracy on average, while MonF$k$NN-PM achieves monotonically reliable predictions. Both have good, stable results in terms of MAE, with AM coming out slightly on top.

AM configuration obtains the most accurate predictions for more than 50\% of the benchmark used. On the other hand, the PM model achieves better results according to monotonicity in 10 of the 12 data-sets used, with large differences in \textit{Windsorhousing} and \textit{MachineCPU} problems. 
When compared with F$k$NN, MonF$k$NN greatly improves the performance of the original algorithm. Both versions of MonF$k$NN (PM and AM) are better on average for each of the three different measures. Particularly, there is an overwhelmingly large difference between F$k$NN and MonF$k$NN-PM in terms of monotonicity. F$k$NN is better only for 3 data-sets when taking just accuracy and MAE into consideration. However, it does not outperform the monotonic predictions of MonF$k$NN.

This improvement is also reflected in the Wilcoxon statistical test applied to the results achieved using these methods. Table \ref{tab:wilcoxom} presents the hypothesis of equivalence of the Wilcoxon test for $\alpha=0.1$ on the pairwise comparison of F$k$NN (1) and our two proposals (MonF$k$NN-PM (2) and MonF$k$NN-AM (3)). As shown in Table \ref{tab:wilcoxom}, MonF$k$NN-AM is statistically better than F$k$NN in terms of accuracy and MAE with $p$-Values under 0.1. Considering monotonicity, MonF$k$NN-PM and -AM statistically outperform F$k$NN with very low $p$-Values. Overall, MonF$k$NN is clearly superior to F$k$NN in scenarios with monotonic constraints.

\begin{table}[h]
\centering
\caption{Wilcoxon test applied to the results obtained by Fuzzy $k$-NN algorithms: F$k$NN (1), MonF$k$NN-PM (2) and MonF$k$NN-AM (3)}
\begin{tabular}{cccccc}
\toprule
 \textbf{Comparison} & \textbf{$R^{+}$} & \textbf{$R^{-}$} & \textbf{Hypothesis ($\alpha=0.1$)} & \textbf{$p$-Value} \\ 
\midrule
\multicolumn{5}{l}{\textit{Accuracy:}}\\
\midrule
 (2) vs. (1) & 49.0 & 17.0 & Not Rejected & 0.1748 \\ 
 (3) vs. (1) & 61.5 & 16.5 & \textbf{Rejected} & 0.0847 \\
\midrule 
\midrule
\multicolumn{5}{l}{\textit{MAE:}}\\ 
\midrule
 (2) vs. (1) & 51.0 & 15.0 & Not Rejected & 0.1230 \\ 
 (3) vs. (1) & 57.0 & 9.0 & \textbf{Rejected} & 0.0322 \\
\midrule 
\midrule
\multicolumn{5}{l}{\textit{NMI:}}\\ 
\midrule
 (2) vs. (1) & 76.5 & 1.50 & \textbf{Rejected} & 0.0012 \\ 
 (3) vs. (1) & 72.5 & 5.50 & \textbf{Rejected} & 0.0059 \\
\bottomrule 
\end{tabular}
\label{tab:wilcoxom}
\end{table}

The reasons for these differences in results are clear and mainly due to their algorithmic differences. MonF$k$NN has learning procedures with notions in the order relation of classes and the monotonic constraints between input and output, which explain an overall better performance in terms of MAE and NMI. Additionally, MonF$k$NN has a greater awareness and treatment of noisy data, which helps obtain better accuracy.

Since monotonicity is usually prioritized in classification with monotonic constraints, we will use MonF$k$NN-PM in the following empirical studies.

\subsection{Comparison with the State-of-the-Art: Prediction capabilities}
\label{subs:exp-State-acc}

Here we evaluate the performance of our approach in comparison to methods from the state-of-the-art of monotonic classification. In this comparison, we look for a balance between accurate and monotonic predictions. Therefore, we compare the results obtained in terms of the selected metrics independently. Then, we draw our conclusions and check if our approach behaves well in the different aspects of classification with monotonic constraints.

First, we evaluate the prediction capability of our method. Table \ref{tab:state-acc} gathers the accuracy results for the different data-sets obtained by the tested algorithms. With these outcomes, MonF$k$NN-PM performs overwhelmingly better than the rest in terms of accuracy. Our approach achieves the most accurate predictions on average with a wide margin. Additionally, it obtains the best results for 5 data-sets, with particularly remarkable cases, such as \textit{balance}. PMDT is the second best method in terms of accuracy and it is the only method that come close to the performance of MonF$k$NN-PM. However, it obtains the overall best results for one data-set only (\textit{bostonhousing}).


\begin{table}[hbt]
  \centering
  \caption{Results in terms of Accuracy achieved by the tested algorithms}
  \resizebox{\textwidth}{!}{
    \begin{tabular}{lrrrrrrrr}
    \toprule
 & \textbf{MonF$k$NN-PM} & \textbf{M$k$NN} & \textbf{OSDL}   & \textbf{OLM} & \textbf{MonMLP} & \textbf{MID}    & \textbf{RDMT} & \textbf{PMDT}   \\
 \midrule
\textit{artiset}                 & 0.9309                & 0.9199          & 0.1952          & 0.7948       & \textbf{0.9463} & 0.7237          & 0.8749        & 0.8539          \\
\textit{balance}                 & \textbf{0.9307}       & 0.8624          & 0.6352          & 0.8320       & 0.9131          & 0.7808          & 0.7216        & 0.7792          \\
\textit{bostonhousing4cl}        & 0.6561                & 0.6126          & 0.2787          & 0.5277       & 0.3979          & 0.6739          & 0.6304        & \textbf{0.6739} \\
\textit{car}                     & \textbf{0.9740}       & 0.9711          & 0.9549          & 0.9543       & 0.8474          & 0.8027          & 0.7297        & 0.9682          \\
\textit{ERA}                     & 0.2420                & 0.1990          & 0.2320          & 0.1690       & 0.2380          & \textbf{0.2760} & 0.2390        & 0.2430          \\
\textit{ESL}                     & 0.7036                & 0.6332          & 0.6721          & 0.5738       & \textbf{0.7234} & 0.6414          & 0.5635        & 0.6598          \\
\textit{LEV}                     & 0.6377                & 0.4630          & \textbf{0.6400} & 0.4250       & 0.6167          & 0.6070          & 0.5210        & 0.6370          \\
\textit{machineCPU}              & \textbf{0.7033}       & 0.6890          & 0.2919          & 0.6746       & 0.6730          & 0.6220          & 0.6555        & 0.6507          \\
\textit{qualitative\_bankruptcy} & \textbf{0.9960}       & \textbf{0.9960} & 0.9160          & 0.9800       & 0.6427          & 0.9840          & 0.9840        & 0.9920          \\
\textit{SWD}                     & 0.5807                & 0.5200          & \textbf{0.5840} & 0.4160       & 0.5063          & 0.5540          & 0.5180        & 0.5830          \\
\textit{windsorhousing}          & 0.7576                & 0.5861          & 0.4927          & 0.7564       & 0.7790          & \textbf{0.8205} & 0.8022        & 0.7564          \\
\textit{wisconsin}               & \textbf{0.9653}       & 0.9649          & 0.9590          & 0.8873       & 0.8604          & 0.9517          & 0.9502        & 0.9561          \\
\midrule
\textit{Avg:}                    & \textbf{0.7565}       & 0.7014          & 0.5710          & 0.6659       & 0.6787          & 0.7031          & 0.6825        & 0.7294        \\ 
\bottomrule
    \end{tabular}%
    }
  \label{tab:state-acc}%
\end{table}%

As mentioned before, we have used the Friedman rank test and the Bayesian Sign test to corroborate the significance of the differences of our approach and the selected methods. Table \ref{tab:friedman-acc} includes the outcome of the Friedman rank and Holm tests in relation to the obtained Accuracy results. MonF$k$NN-PM is ranked first with a high ranking value compared to others. All the hypotheses of equivalence are rejected with small $p$-values with the exception of PMDT, which would be rejected for $\alpha = 0.1$. The distance between the ranks of MonF$k$NN-PM and PMDT is still quite large. 

\begin{table}[ht]
  \centering
  \caption{Holm test applied to the Accuracy results among the tested algorithms}
  \resizebox{0.8\textwidth}{!}{
    \begin{tabular}{lllll}
    \toprule
 \multicolumn{4}{c}{\textbf{Control Method:} MonF$k$NN-PM (2.04)}  &  \\
    \midrule
 \textbf{i} & \textbf{Algorithm (Rank)} & \textbf{Z} & \textbf{$p$-Value} & \textbf{Hypothesis ($\alpha = 0.05$)}\\
    \midrule
7&OLM (6.13) &4.083&0.00004& \textbf{Rejected}\\
6&OSDL (5.42) &3.375&0.00073& \textbf{Rejected}\\
5&RDMT (5.38) &3.333&0.00085& \textbf{Rejected}\\
4&MonMLP (4.67) &2.625&0.00866& \textbf{Rejected}\\
3&MID (4.42) &2.375&0.01754& \textbf{Rejected}\\
2&M$k$NN (4.21) &2.167&0.03026& \textbf{Rejected}\\
1&PMDT (3.75) &1.708&0.08757& Not Rejected \\

    \bottomrule
    \end{tabular}%
    }
  \label{tab:friedman-acc}%
\end{table}%

\begin{figure}[htb]
 	\begin{center}
     \subfloat[vs. M$k$NN\label{fig:bayacc-mknn}]{\includegraphics[trim={1.7cm 2.5cm 1.7cm 2.5cm},clip,scale=0.25]{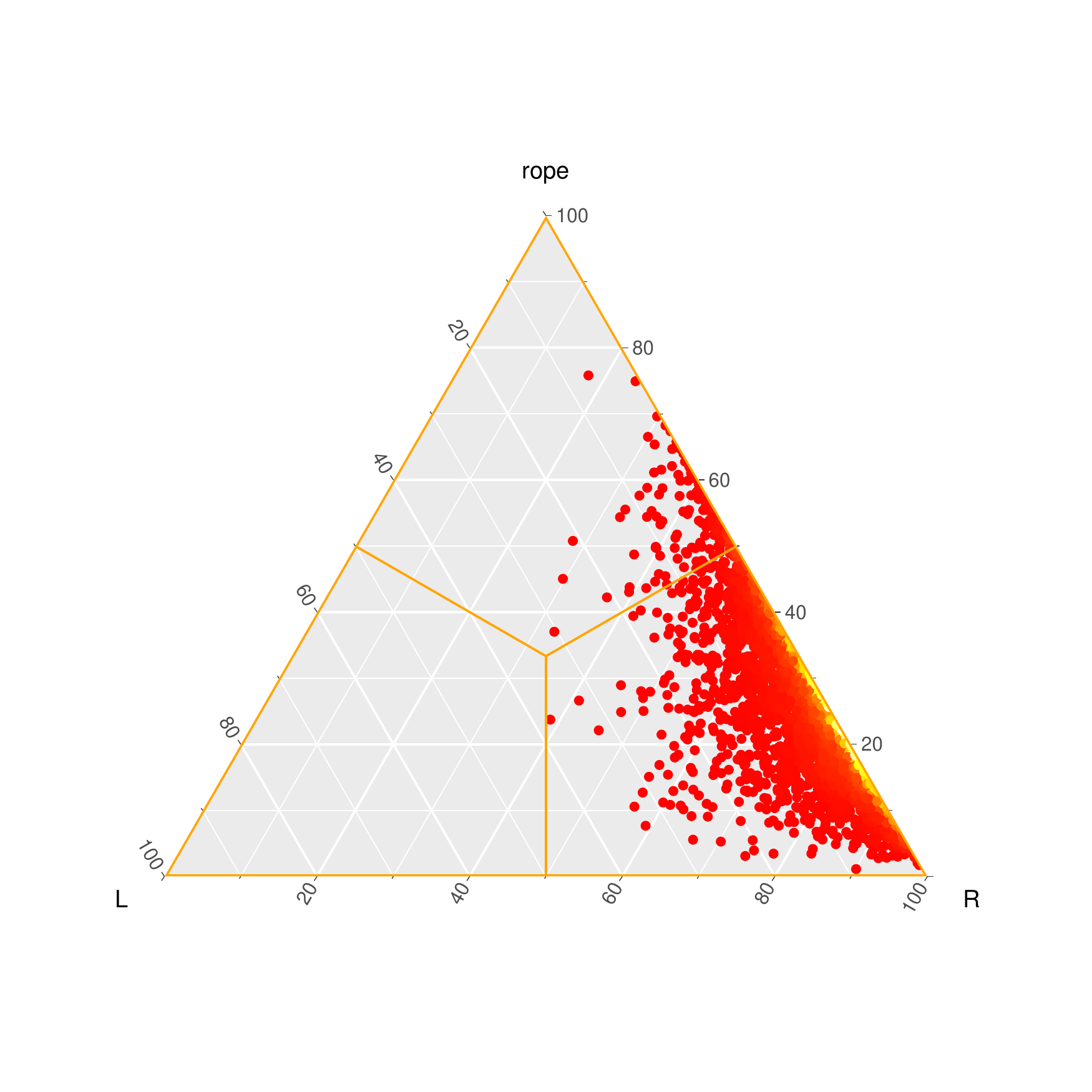}}
 	\subfloat[vs. OSDL \label{fig:bayacc-osdl}]{\includegraphics[trim={1.7cm 2.5cm 1.7cm 2.5cm},clip,scale=0.25]{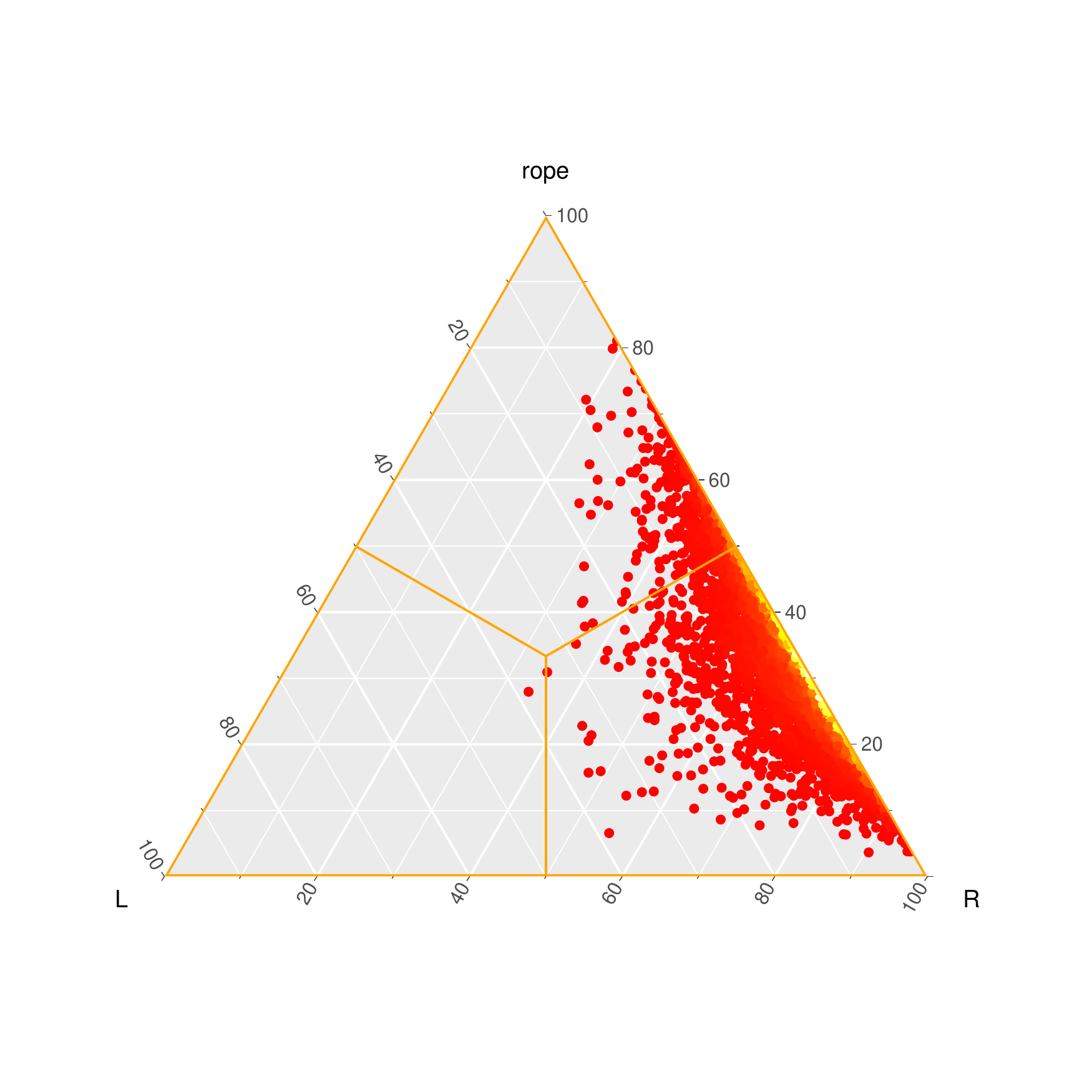}}
  \subfloat[vs. OLM\label{fig:bayacc-olm}]{\includegraphics[trim={1.7cm 2.5cm 1.7cm 2.5cm},clip,scale=0.25]{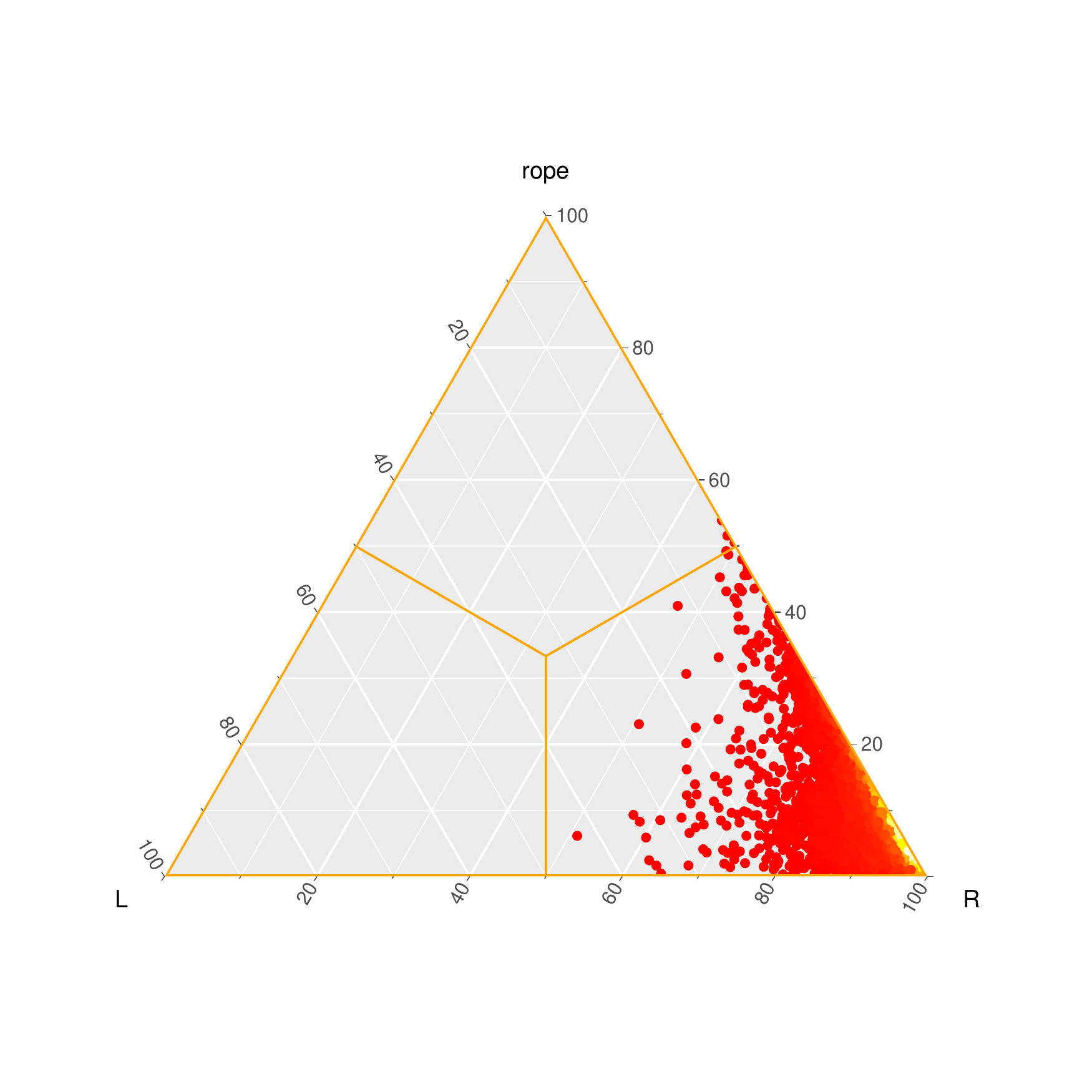}}
  
 \subfloat[vs. MonMLP\label{fig:bayacc-mlp}]{\includegraphics[trim={1.7cm 2.5cm 1.7cm 2.5cm},clip,scale=0.25]{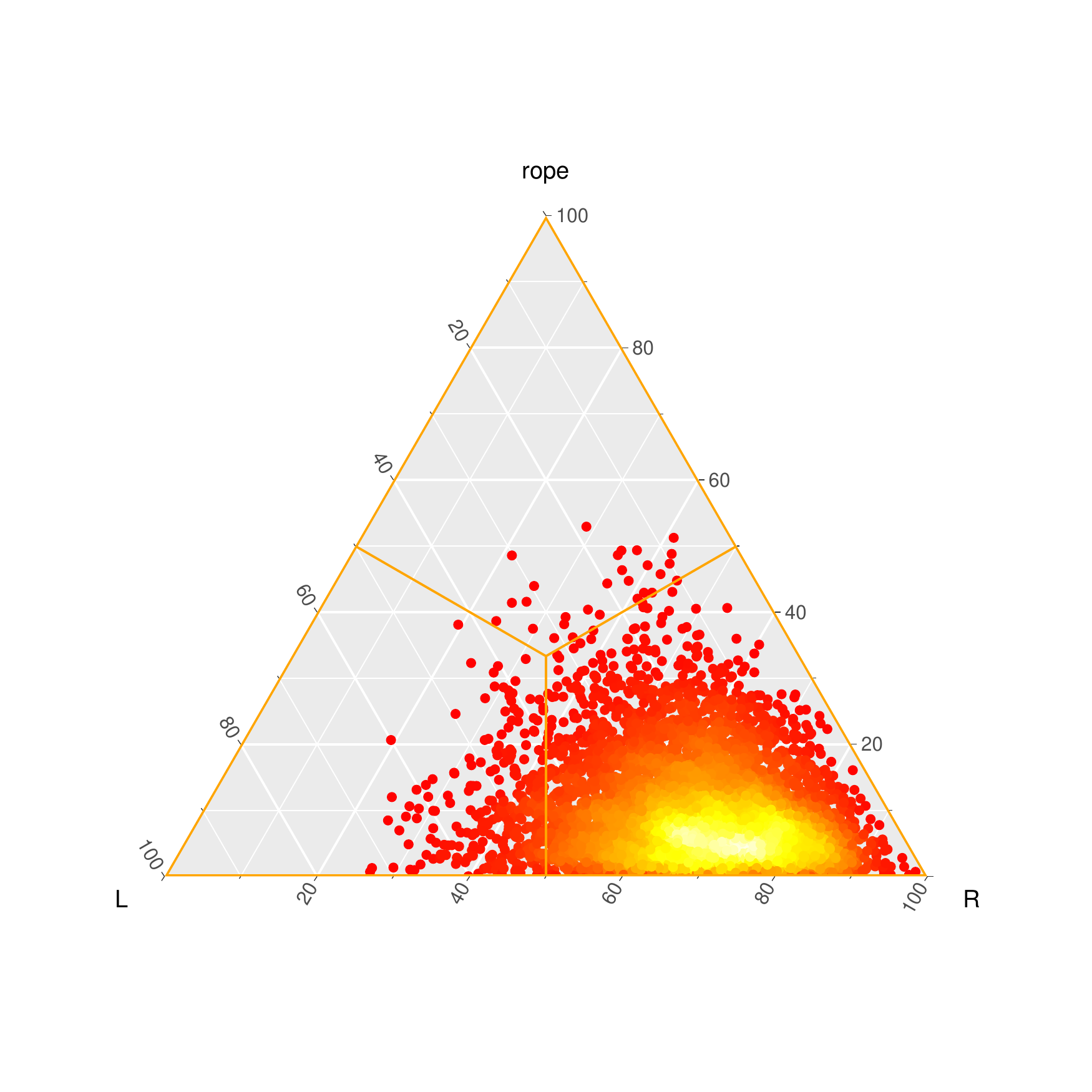}}
\subfloat[vs. PMDT\label{fig:bayacc-pmdt}]{\includegraphics[trim={1.7cm 2.5cm 1.7cm 2.5cm},clip,scale=0.25]{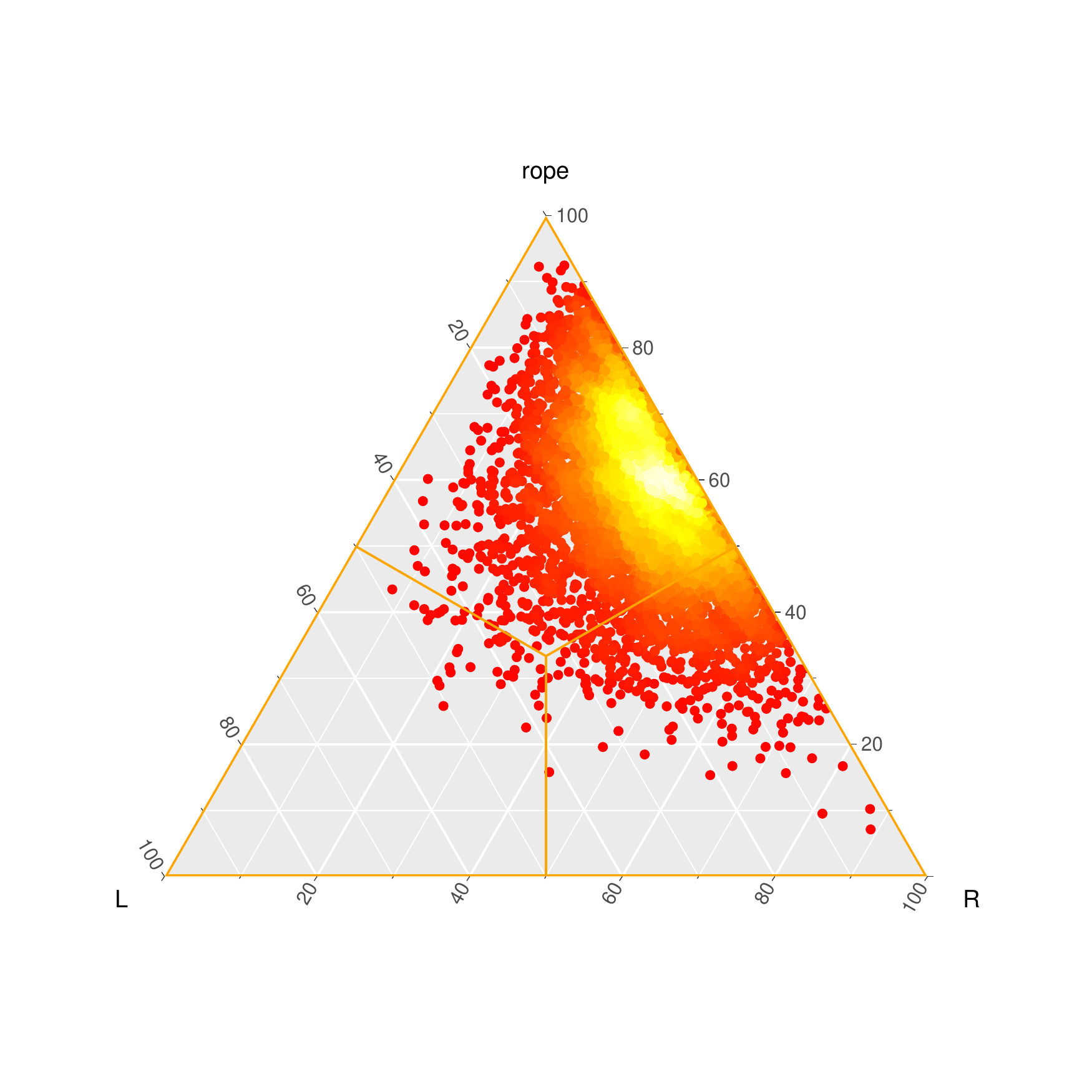}}
 		\caption{Bayesian Sign Test heat-map for MonF$k$NN-PM vs. the rest in terms of accuracy.}
		\label{fig:bayes-acc}		
	\end{center}
 \end{figure}

Figure \ref{fig:bayes-acc} graphically represents the difference between MonF$k$NN-PM and other methods and its statistical significance in terms of accuracy. In order to save space and avoid plotting 7 heat-maps for each metric, we have only included PMDT, as it is the best and most recent algorithm among the monotonic decision trees \cite{pei18}. As mentioned before, the position of the majority of the distribution in these maps determines the decision of the test: the right sector means the statistical superiority of MonF$k$NN-PM over the compared method, the rope sector is the statistical equivalency and the left side indicates the superiority of the other algorithm.

These heat-maps clearly indicate the significant superiority of MonF$k$NN-PM over all methods except PMDT as the computed distributions are always located in the right region. The most significant outcome is the comparison with OLM (Figure \ref{fig:bayacc-olm}), even though it does not obtain the worst results. For M$k$NN (Figure \ref{fig:bayacc-mknn}) and OSDL (Figure \ref{fig:bayacc-osdl}), there are a few cases where their performances are statically equivalent to MonF$k$NN-PM. On the contrary, MonMLP is significantly more accurate in a few data-sets, although the MonF$k$NN-PM is clearly superior (Figure \ref{fig:bayacc-mlp}). Considering the comparison with PMDT (Figure \ref{fig:bayacc-mlp}), the majority of the distribution is located in the statistical equivalence. However, it is still shifted to the right with a large number of points, indicating a better performance for MonF$k$NN-PM. Almost none support the performance of PMDT.  

Error costs could be essential for monotonic ranking problems. Table \ref{tab:state-mae} shows the error in the form of MAE made by the evaluated classifiers. As was the case in accuracy performance,  MonF$k$NN-PM clearly performs better than the rest, with the smallest error on average and for 4 of the data-sets. It also achieves similar results in problems where other algorithms come out on top, such as \textit{LEV} or \textit{wisconsin}.



\begin{table}[ht]
  \centering
  \caption{Results in terms of MAE achieved by the tested algorithms}
  \resizebox{\textwidth}{!}{
    \begin{tabular}{lrrrrrrrr}
    \toprule
                                 & \textbf{MonF$k$NN-PM} & \textbf{M$k$NN} & \textbf{OSDL}   & \textbf{OLM} & \textbf{MonMLP} & \textbf{MID}    & \textbf{RDMT} & \textbf{PMDT}   \\
                                 \midrule
\textit{artiset}                 & 0.0691                & 0.0771          & 1.6897          & 0.2082       & \textbf{0.0537} & 0.3123          & 0.1251        & 0.1471          \\
\textit{balance}                 & \textbf{0.0853}       & 0.1504          & 0.4912          & 0.1920       & 0.0992          & 0.3360          & 0.3840        & 0.2560          \\
\textit{bostonhousing4cl}        & 0.3972                & 0.4901          & 0.9368          & 0.5988       & 0.7655          & 0.3893          & 0.4249        & \textbf{0.3676} \\
\textit{car}                     & \textbf{0.0295}       & 0.0359          & 0.0475          & 0.0538       & 0.1599          & 0.2506          & 0.3079        & 0.0365          \\
\textit{ERA}                     & 1.2813                & 1.4270          & 1.2850          & 2.1500       & \textbf{1.2317} & 1.2970          & 1.3060        & 1.2870          \\
\textit{ESL}                     & 0.3149                & 0.3791          & 0.3607          & 0.4734       & \textbf{0.2910} & 0.3934          & 0.4918        & 0.3750          \\
\textit{LEV}                     & 0.3927                & 0.5740          & \textbf{0.3920} & 0.6680       & 0.4170          & 0.4290          & 0.5430        & 0.3940          \\
\textit{machineCPU}              & \textbf{0.3158}       & 0.3301          & 0.9043          & 0.3589       & 0.3413          & 0.4211          & 0.3589        & 0.3732          \\
\textit{qualitative\_bankruptcy} & \textbf{0.0040}       & \textbf{0.0040} & 0.0840          & 0.0200       & 0.3573          & 0.0160          & 0.0160        & 0.0080          \\
\textit{SWD}                     & 0.4370                & 0.4840          & 0.4370          & 0.7630       & 0.5167          & 0.4750          & 0.4990        & \textbf{0.4340} \\
\textit{windsorhousing}          & 0.2424                & 0.4304          & 0.5073          & 0.2436       & 0.2210          & \textbf{0.1795} & 0.1978        & 0.2436          \\
\textit{wisconsin}               & 0.0347                & \textbf{0.0337} & 0.0410          & 0.1127       & 0.1396          & 0.0483          & 0.0498        & 0.0439          \\
\midrule
\textit{Avg:}                    & \textbf{0.3003}       & 0.3680          & 0.5980          & 0.4869       & 0.3828          & 0.3790          & 0.3920        & 0.3305        \\ 
\bottomrule
    \end{tabular}%
    }
  \label{tab:state-mae}%
\end{table}%

Table \ref{tab:friedman-mae} shows the ranking of the methods and $p$-values obtained with the post hoc test for the MAE comparison. As in the accuracy tests, our proposal is once again ranked as the best method with a solid statistical significance as compared to almost all algorithms. PMDT still achieves similar results to MonF$k$NN-PM with a $p$-value that does not reject the hypothesis for $\alpha = 0.05$, but does for $\alpha = 0.1$. In this case, the $p$-value of PMDT is smaller and its rank difference with our proposal is larger than that obtained in terms of accuracy.

\begin{table}[ht]
  \centering
  \caption{Holm test applied to the MAE results among the tested algorithms}
  \resizebox{0.8\textwidth}{!}{
    \begin{tabular}{lllll}
    \toprule
 \multicolumn{4}{c}{\textbf{Control Method:} MonF$k$NN-PM (2.00)}  &  \\
    \midrule
 \textbf{i} & \textbf{Algorithm (Rank)} & \textbf{Z} & \textbf{$p$-Value} & \textbf{Hypothesis ($\alpha = 0.05$)}\\
    \midrule
    7&OLM (6.17) &4.167&0.00003&\textbf{Rejected}\\
    6&RDMT (5.54) &3.542&0.00040&\textbf{Rejected}\\
    5&OSDL (5.29) &3.292&0.00099&\textbf{Rejected}\\
    4&MID (4.96) &2.958&0.00309&\textbf{Rejected}\\
    3&MonMLP (4.25) &2.250&0.02445&\textbf{Rejected}\\
    2&M$k$NN (4.04) &2.042&0.04119&\textbf{Rejected}\\
    1&PMDT (3.75) &1.750&0.08011&Not Rejected\\
    \bottomrule
    \end{tabular}%
    }
  \label{tab:friedman-mae}%
\end{table}%

\begin{figure}[ht]
 	\begin{center}
     \subfloat[vs. M$k$NN\label{fig:baymae-mknn}]{\includegraphics[trim={1.7cm 2.5cm 1.7cm 2.5cm},clip,scale=0.25]{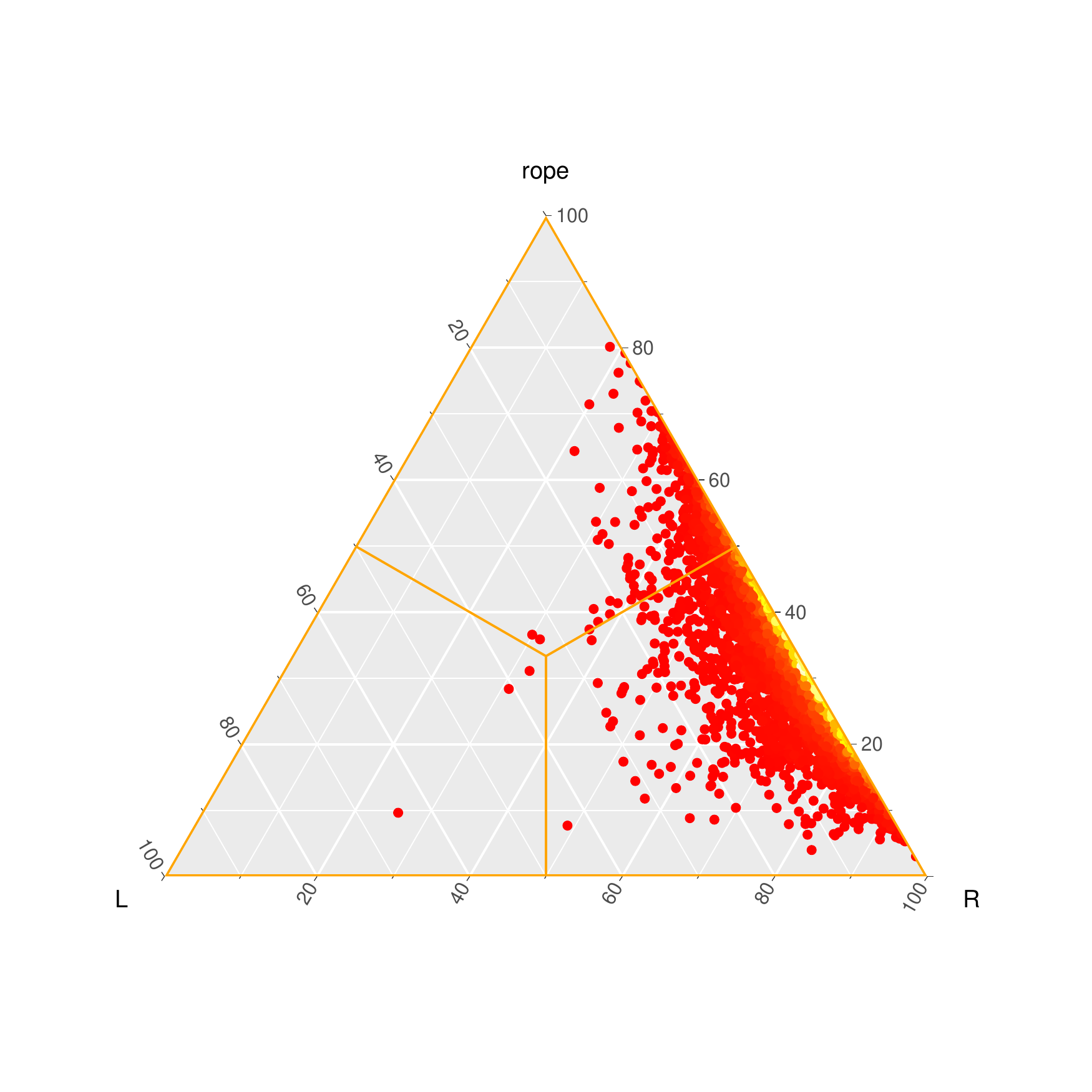}}
 	\subfloat[vs. OSDL \label{fig:baymae-osdl}]{\includegraphics[trim={1.7cm 2.5cm 1.7cm 2.5cm},clip,scale=0.25]{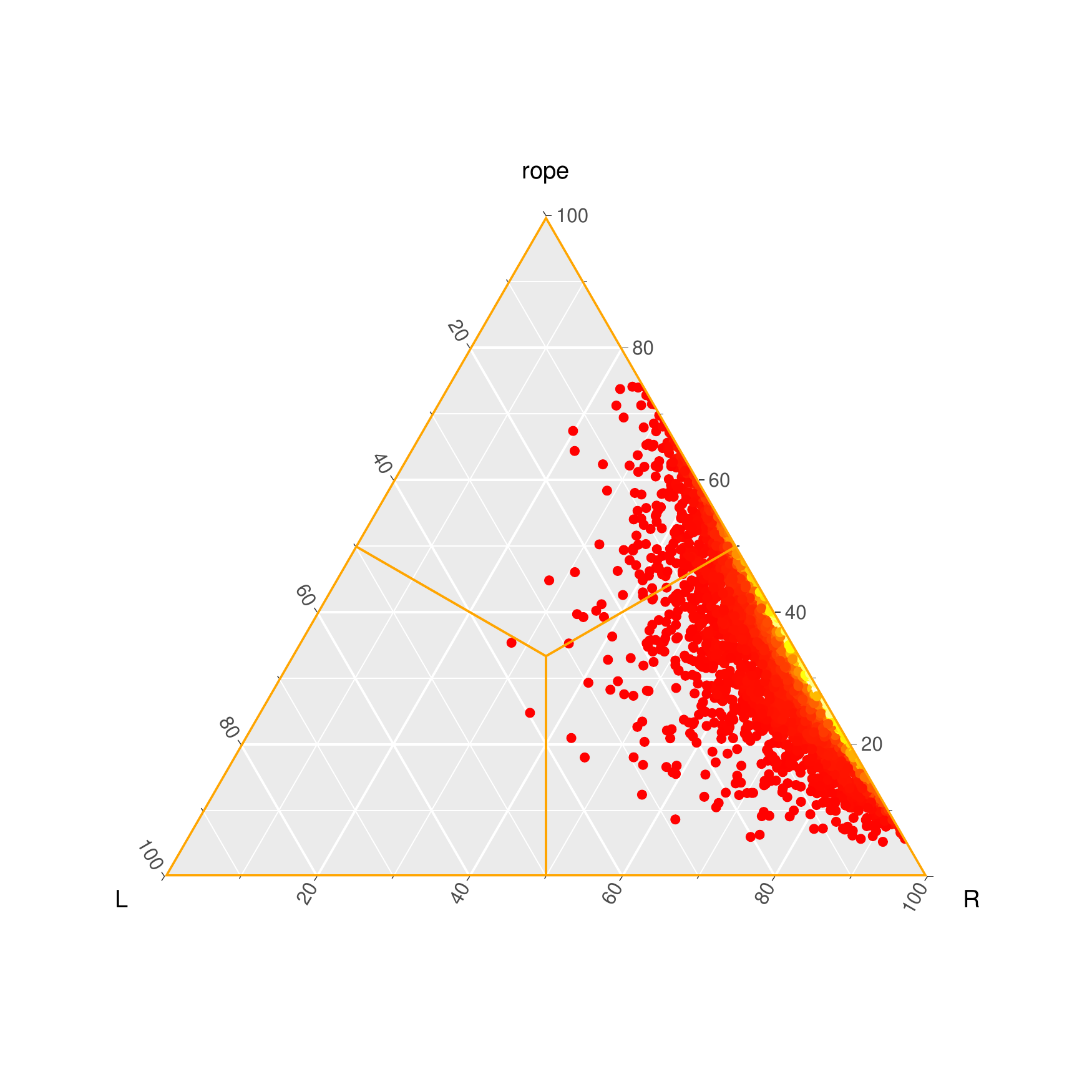}}
  \subfloat[vs. OLM\label{fig:baymae-olm}]{\includegraphics[trim={1.7cm 2.5cm 1.7cm 2.5cm},clip,scale=0.25]{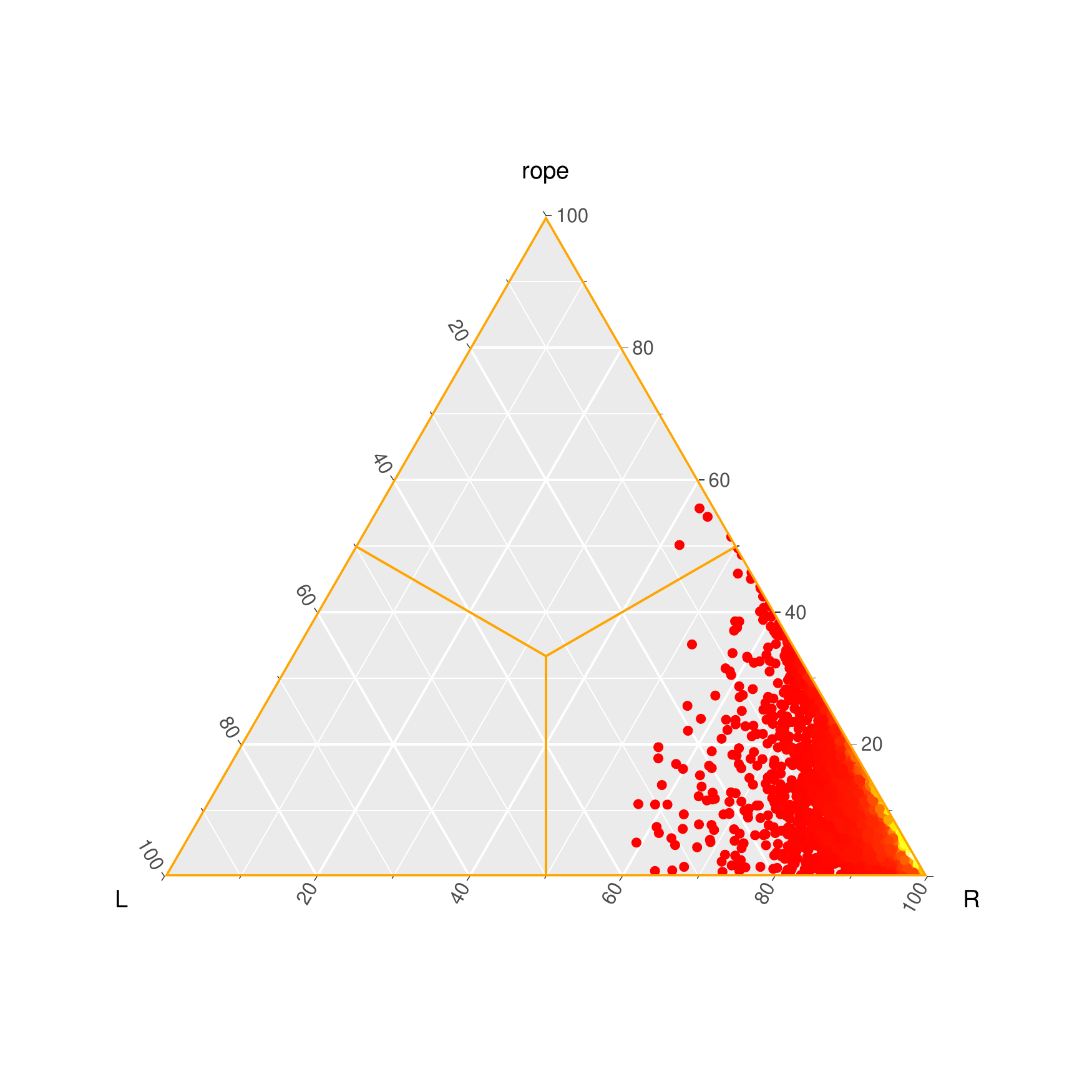}}
  
 \subfloat[vs. MonMLP\label{fig:baymae-mlp}]{\includegraphics[trim={1.7cm 2.5cm 1.7cm 2.5cm},clip,scale=0.25]{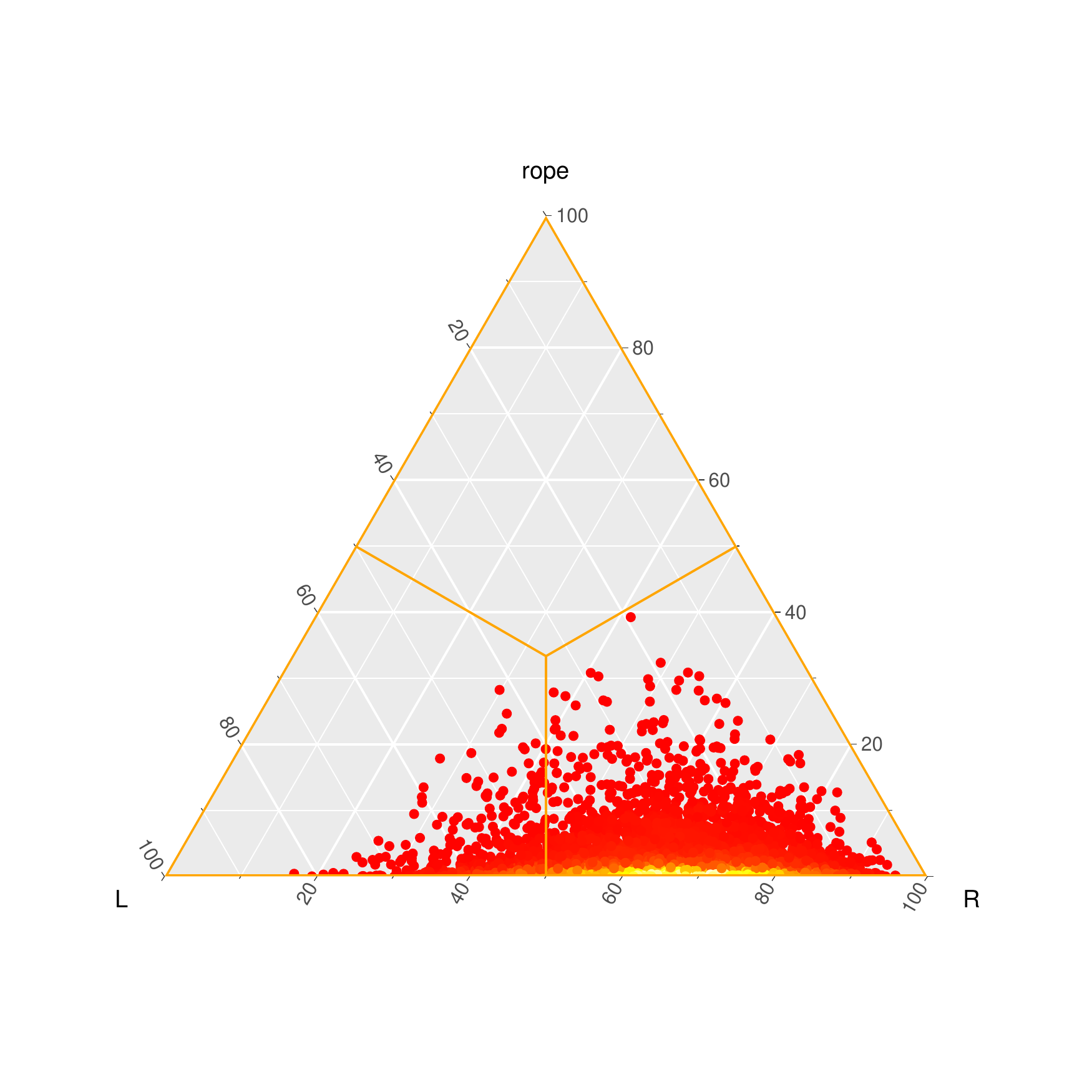}}
\subfloat[vs. PMDT\label{fig:baymae-pmdt}]{\includegraphics[trim={1.7cm 2.5cm 1.7cm 2.5cm},clip,scale=0.25]{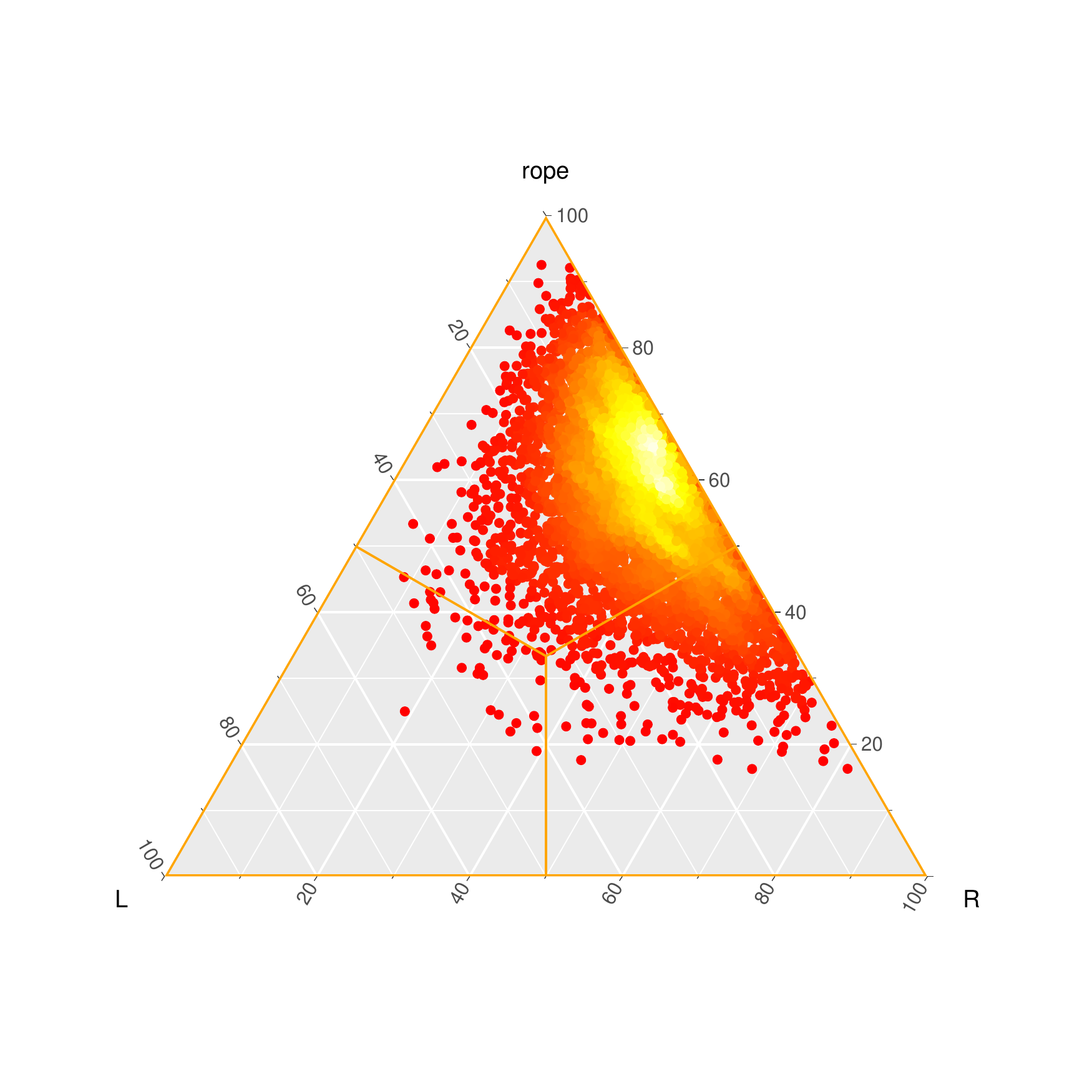}}
 		\caption{Bayesian Sign Test heat-map for MonF$k$NN-PM vs. the rest in terms of MAE.}
		\label{fig:bayes-mae}		
	\end{center}
 \end{figure}

Figure \ref{fig:bayes-mae} shows the Bayesian Sign test on pairwise comparison with our method according to MAE. As shown by the distributions in the right part of the majority of the figures, MonF$k$NN-PM is definitely better when considering error costs. This is more statistically significant as compared to OLM (Figure \ref{fig:baymae-olm}), where nearly the entire distribution is in the right region. MonF$k$NN, M$k$NN and OSDL share some good results, but these last two are not statistically better than the former in any circumstance as seen in Figure \ref{fig:baymae-mknn} and Figure \ref{fig:baymae-osdl}. As we have also seen in the accuracy comparison, Figure \ref{fig:baymae-mlp} points out the statistical superiority of MonF$k$NN-PM over MonMLP, but the latter has a better MAE in some cases. Given Figure \ref{fig:baymae-pmdt}, MonF$k$NN-PM and PMDT can be considered to be statistically the same in terms of error costs. However, MonF$k$NN-PM performs better statistically than PMDT in an important part of the benchmark, as a fragment of the distribution is located on the right side and almost none are found on the left.

\subsection{Comparison with the State-of-the-Art: Monotonicity}
\label{subs:exp-State-mon}

Now we will analyze the performance according to the monotonicity of our proposal compared to methods chosen from the state-of-the-art. Table \ref{tab:state-nmi} shows the NMI results achieved by the selected models. In this case, the competition is close. Monotonic decision trees (MID, RDMT, and PMDT) clearly obtain less monotonic predictions. MID has the worst behavior considering only monotonicity and PMDT is the most monotonic decision tree classifier. OLM and MonMLP are slightly better than PMDT, but they still do not come close to the best methods. MonF$k$NN-PM, M$k$NN, and OSDL perform similarly. MonF$k$NN-PM and OSDL are slightly better on average. It is worth mentioning the existence of simpler data-sets, such as \textit{artiset} and \textit{wisconsin}, in relation to monotonicity as almost every algorithm accomplishes the same good results. The best results for the more complex sets are shared by the different methods.


\begin{table}[ht]
  \centering
  \caption{Results in terms of NMI achieved by the tested algorithms}
  \resizebox{\textwidth}{!}{
    \begin{tabular}{lrrrrrrrr}
    \toprule
    & \textbf{MonF$k$NN-PM} & \textbf{M$k$NN} & \textbf{OSDL}   & \textbf{OLM}    & \textbf{MonMLP} & \textbf{MID}    & \textbf{RDMT}   & \textbf{PMDT}   \\
                            \midrule
\textit{artiset}                 & \textbf{0.0000}       & \textbf{0.0000} & \textbf{0.0000} & \textbf{0.0000} & \textbf{0.0000} & 0.0039          & \textbf{0.0000} & 0.0001          \\
\textit{balance}                 & \textbf{0.0000}       & 0.0001          & 0.0006          & \textbf{0.0000} & \textbf{0.0000} & 0.0017          & 0.0029          & 0.0010          \\
\textit{bostonhousing4cl}        & \textbf{0.0000}       & \textbf{0.0000} & \textbf{0.0000} & 0.0003          & 0.0007          & 0.0022          & 0.0010          & 0.0010          \\
\textit{car}                     & \textbf{0.0000}       & \textbf{0.0000} & \textbf{0.0000} & \textbf{0.0000} & 0.0001          & 0.0046          & 0.0002          & \textbf{0.0000} \\
\textit{ERA}                     & 0.0052                & 0.0056          & 0.0049          & 0.0063          & \textbf{0.0026} & 0.0082          & 0.0085          & 0.0058          \\
\textit{ESL}                     & 0.0004                & 0.0012          & 0.0006          & 0.0025          & \textbf{0.0003} & 0.0021          & 0.0066          & 0.0032          \\
\textit{LEV}                     & \textbf{0.0004}       & 0.0010          & \textbf{0.0004} & 0.0043          & 0.0008          & 0.0018          & 0.0086          & 0.0006          \\
\textit{machineCPU}              & 0.0002                & \textbf{0.0000} & \textbf{0.0000} & 0.0014          & 0.0001          & 0.0037          & 0.0047          & 0.0028          \\
\textit{qualitative\_bankruptcy} & \textbf{0.0000}       & \textbf{0.0000} & 0.0003          & \textbf{0.0000} & 0.0079          & 0.0002          & \textbf{0.0000} & \textbf{0.0000} \\
\textit{SWD}                     & 0.0007                & 0.0005          & 0.0009          & 0.0015          & 0.0004          & 0.0020          & \textbf{0.0000} & 0.0010          \\
\textit{windsorhousing}          & 0.0005                & \textbf{0.0000} & \textbf{0.0000} & \textbf{0.0000} & \textbf{0.0000} & 0.0030          & 0.0002          & 0.0059          \\
\textit{wisconsin}               & \textbf{0.0000}       & \textbf{0.0000} & \textbf{0.0000} & \textbf{0.0000} & \textbf{0.0000} & \textbf{0.0000} & 0.0001          & \textbf{0.0000} \\
\midrule
 \textit{Avg:}                    & \textbf{0.0006}       & 0.0007          & \textbf{0.0006} & 0.0014          & 0.0011          & 0.0028          & 0.0027          & 0.0018         \\
\bottomrule
    \end{tabular}%
    }
  \label{tab:state-nmi}%
\end{table}%

Table \ref{tab:friedman-nmi} summarizes the comparison according to monotonicity with the Friedman statistical test results. In this case, MonF$k$NN-PM is barely selected as the control method. For half of the benchmark (OSDL, M$k$NN, MonMLP and OLM), the hypotheses of equivalence are not rejected for $\alpha = 0.05$. On the contrary, all monotonic decision trees are statistically worse than MonF$k$NN-PM by a wide margin. The best monotonic decision tree (PMDT) does not reach good performance in terms of monotonicity of the best algorithms. This is probably due to the greedy construction of monotonic constraints into the tree.  

\begin{table}[ht]
  \centering
  \caption{Holm test applied to the NMI results among the tested algorithms}
  \resizebox{0.8\textwidth}{!}{
    \begin{tabular}{lllll}
    \toprule
 \multicolumn{4}{c}{\textbf{Control Method:} MonF$k$NN-PM (2.9583)}  &  \\
    \midrule
 \textbf{i} & \textbf{Algorithm (Rank)} & \textbf{Z} & \textbf{$p$-Value} & \textbf{Hypothesis ($\alpha = 0.05$)}\\
    \midrule
    
    7&MID (7.00) &4.042&0.00005&\textbf{Rejected}\\
    6&RDMT (6.33) &3.375&0.00074&\textbf{Rejected}\\
    5&PMDT (5.75) &2.792&0.00524&\textbf{Rejected}\\
    4&OLM (4.13) &1.167&0.24335&Not Rejected\\
    3&MonMLP (3.63) &0.667&0.50499&Not Rejected\\
    2&M$k$NN (3.13) &0.167&0.86763&Not Rejected\\
    1&OSDL (3.08) &0.125&0.90052&Not Rejected\\
    
    \bottomrule
    \end{tabular}%
    }
  \label{tab:friedman-nmi}%
\end{table}%

\begin{figure}[ht]
 	\begin{center}
     \subfloat[vs. M$k$NN\label{fig:baynmi-mknn}]{\includegraphics[trim={1.7cm 2.5cm 1.7cm 2.5cm},clip,scale=0.25]{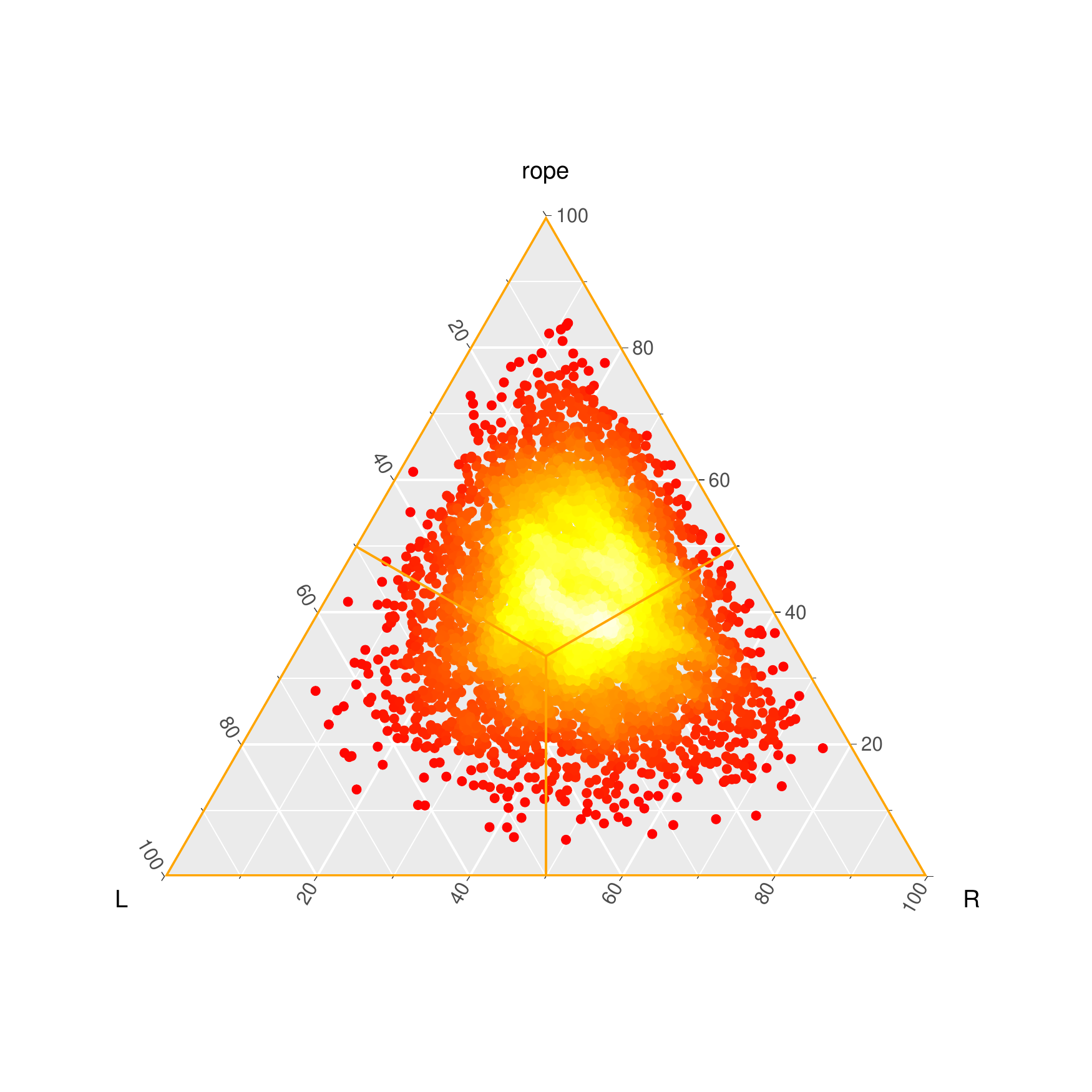}}
 	\subfloat[vs. OSDL \label{fig:baynmi-osdl}]{\includegraphics[trim={1.7cm 2.5cm 1.7cm 2.5cm},clip,scale=0.25]{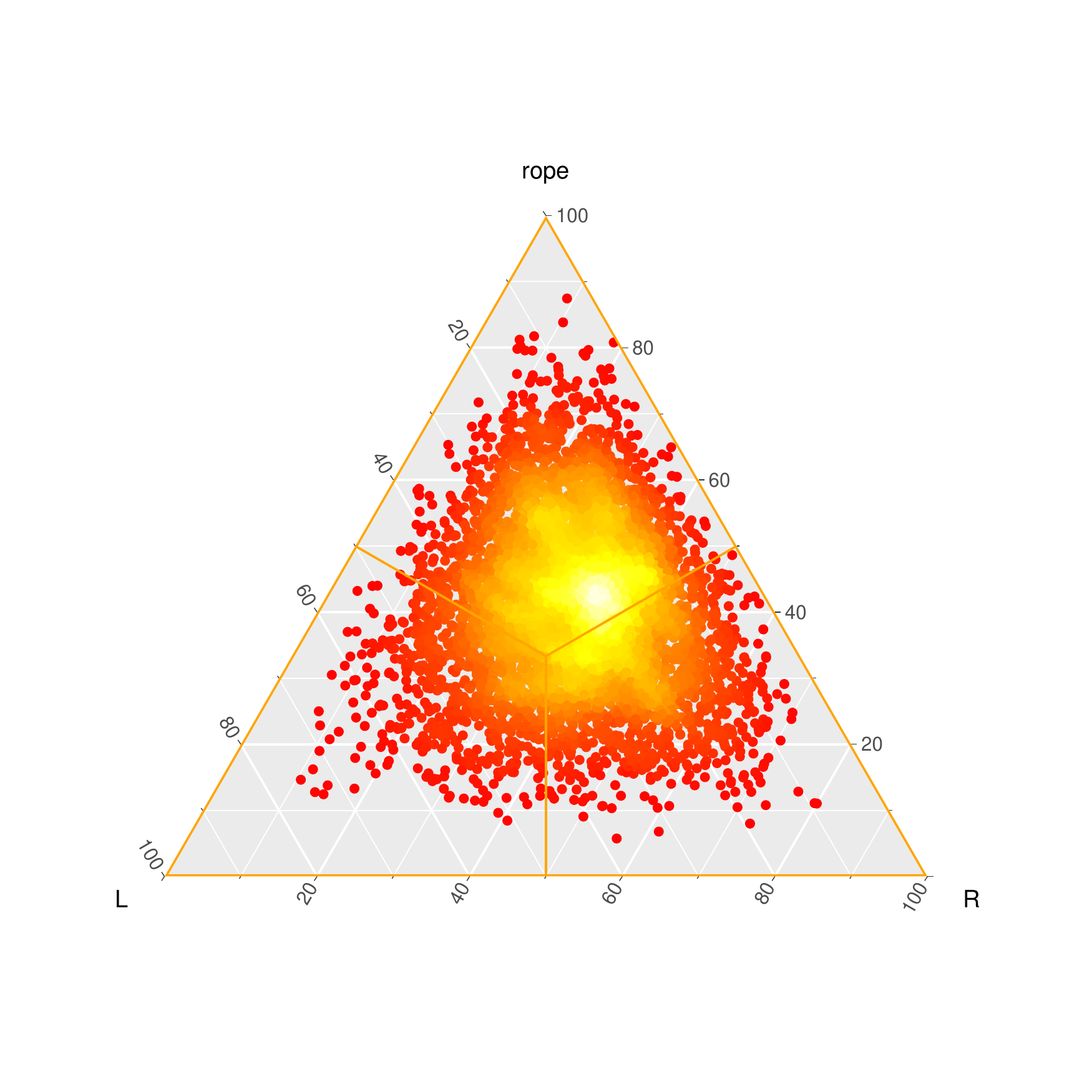}}
  \subfloat[vs. OLM\label{fig:baynmi-olm}]{\includegraphics[trim={1.7cm 2.5cm 1.7cm 2.5cm},clip,scale=0.25]{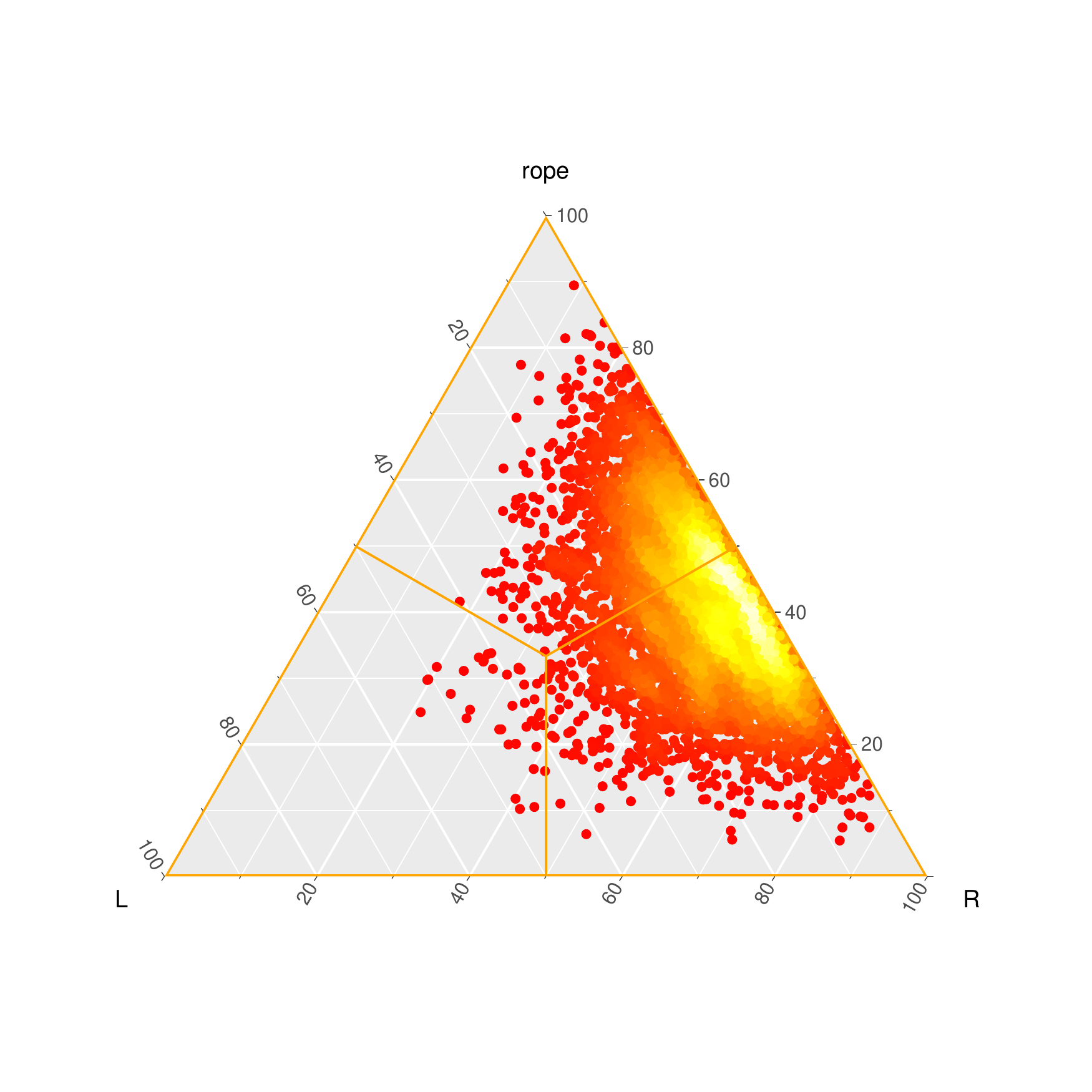}}
  
 \subfloat[vs. MonMLP\label{fig:baynmi-mlp}]{\includegraphics[trim={1.7cm 2.5cm 1.7cm 2.5cm},clip,scale=0.25]{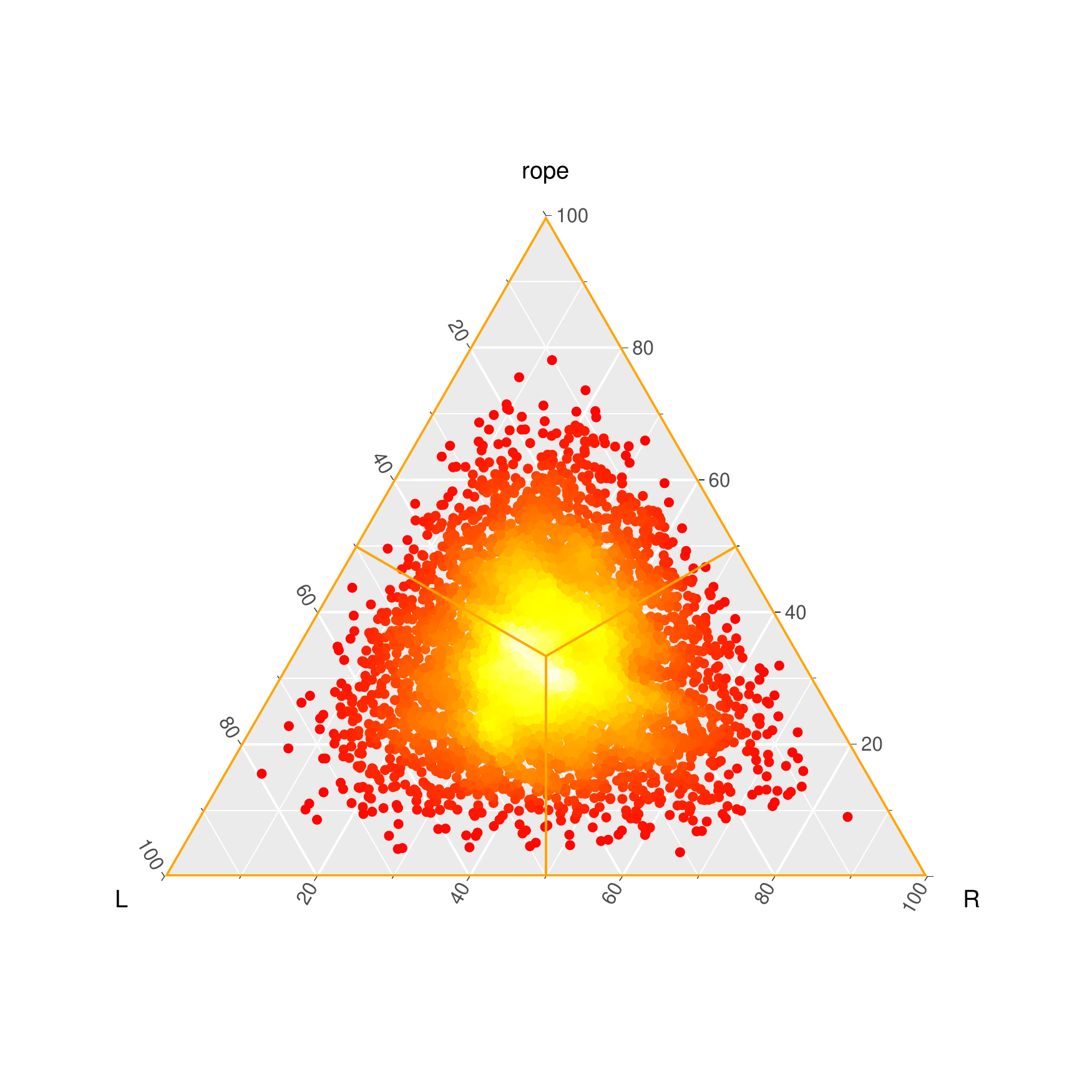}}
\subfloat[vs. PMDT\label{fig:baynmi-pmdt}]{\includegraphics[trim={1.7cm 2.5cm 1.7cm 2.5cm},clip,scale=0.25]{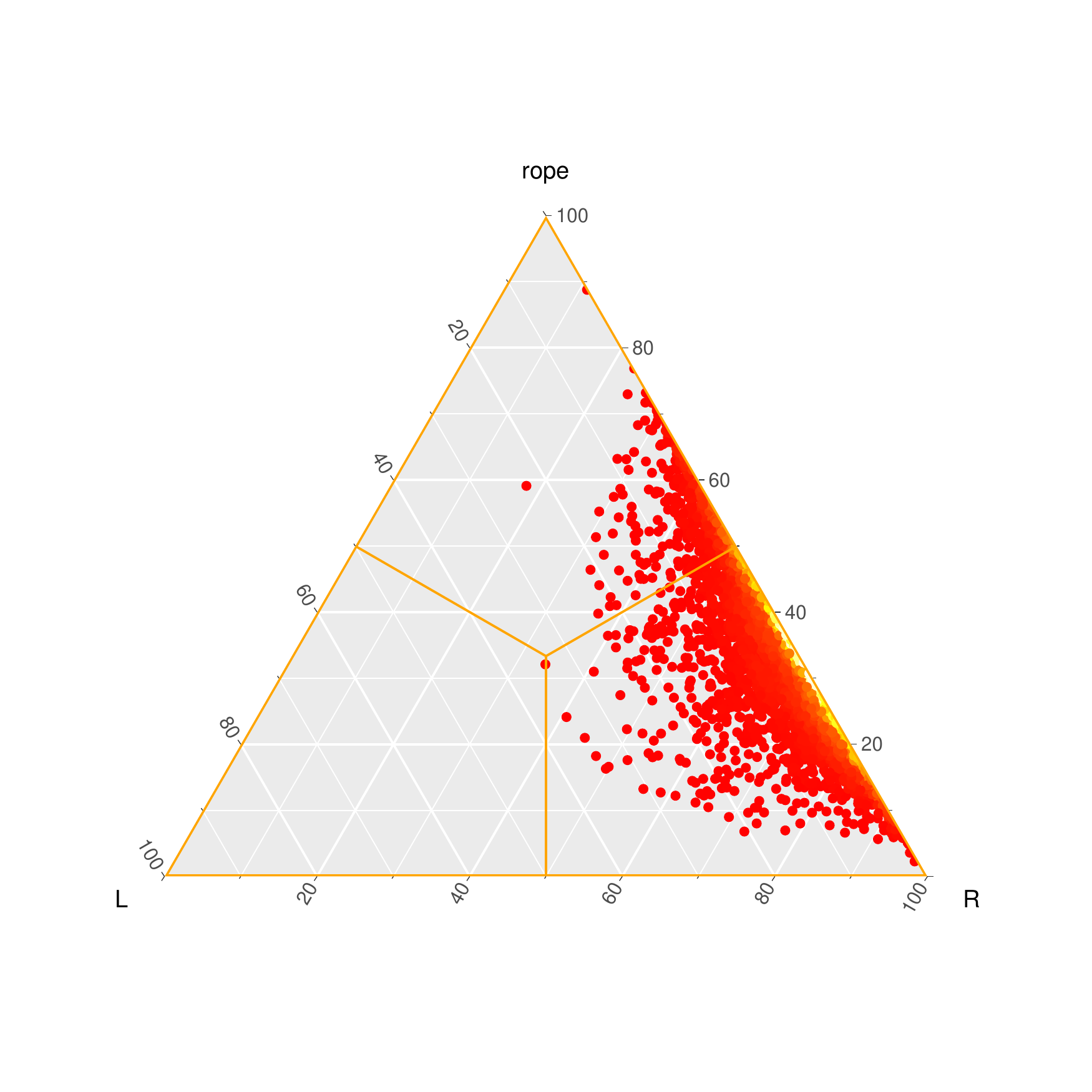}}
 		\caption{Bayesian Sign Test heat-map for MonF$k$NN-PM vs. the rest in terms of NMI.}
		\label{fig:bayes-nmi}		
	\end{center}
 \end{figure}

In Figure \ref{fig:bayes-nmi}, the statistical comparisons of the NMI results are represented with Bayesian Sign Test heat-maps. These plots show similar conclusions extracted from the previous table with NMI results. MonF$k$NN is significantly superior to PMDT (Figure \ref{fig:baynmi-pmdt}). In Figure \ref{fig:baynmi-olm}, the right-shifted distribution points out that MonF$k$NN-PM is better than OLM. Although they share a part of the distribution in the rope section, OLM has too few individuals in its left section (Figure \ref{fig:baynmi-olm}). When compared with M$k$NN (Figure \ref{fig:baynmi-mknn}), OSDL (Figure \ref{fig:baynmi-osdl}) and MonMLP (Figure \ref{fig:baynmi-mlp}), big parts of the distributions are located in all the decision sectors. Even though their distributions are slightly shifted to the right (Figure \ref{fig:baynmi-mknn} and Figure \ref{fig:baynmi-osdl}), the core of the distributions are found in the rope. Then, we can roughly assume statistical equivalence.

In summary, MonF$k$NN-PM obtains significantly better results in terms of accuracy and error cost than almost all of the considered methods. Our approach also achieves the most monotonic predictions alongside OSDL. MonF$k$NN-PM is slightly and non-statistically better than PMDT in terms of accuracy and error costs, but the former overwhelmingly outperforms PMDT considering monotonicity. Therefore, MonF$k$NN-PM is an overall better method. 

The main reason behind the remarkable performance of MonF$k$NN is its capability of not sacrificing any objective of monotonic classification. Usually, some classifiers, such as OSDL, sacrifice accurate predictions in order to accomplish monotonic models. The results of OSDL for \textit{artiset} and \textit{bostonhousing} and the outcome of M$k$NN for \textit{balance} are good examples of this statement. On the other hand, other methods, such as monotonic decision trees and particularly PMDT, achieve accurate predictions but break the monotonic constraints in their predictions more frequently. However, the MonF$k$NN procedure of training class membership extraction is designed to mitigate the influence of non-monotonic noisy data, without the need to aggressively modify the training data as done by relabeling in M$k$NN. The MonF$k$NN prediction stage offers the flexibility of choice for most accurate or monotonic predictions. Additionally, MonF$k$NN includes technologies that are more appropriate for ordinal and monotonic classification, such as median as a final class.


\subsection{On the robustness of Monotonic Fuzzy k-NN to monotonic noise}
\label{subs:exp-noise}

With this last empirical study, we aim to test the robustness of MonF$k$NN-PM to the presence of monotonic violations or noise in the training sets as compared to M$k$NN. Thus, we have introduced different amounts of noisy instances in the training partitions of the artificial data-set \textit{Artiset}. Then, the performance of MonF$k$NN-PM and M$k$NN is measured and compared in terms of accuracy, MAE and NMI while the noise ratio increases.

In order to increase the impact of class noise, we have randomly under-sampled every training set to 25\% of their instances. Then, a subset of randomly selected instances is converted to noise by changing their class labels. This label modification is done according to the adjacent classes of the implicated instance. Specifically, a large number of neighbors are computed for the future noisy example $x_i$. 15 nearest neighbors were the value used in this experiment. Next, the neighbors with the same class as $x_i$ are removed and a new class is randomly obtained in relation to the presence ratio of other classes in its filtered neighbors. This ensures a certain degree of proximity between the changed sample and its new class.

This process is executed following the same cross-validation scheme mentioned earlier. Since the noise generation has a random component, the experiment was repeated three times with different seeds, averaging the obtained results.  After the noise generation and before the execution of M$k$NN, a relabeling technique \cite{feelders10} was applied to the resultant data-sets.

\begin{figure}[t]
 	\begin{center}
     \includegraphics[scale=0.45]{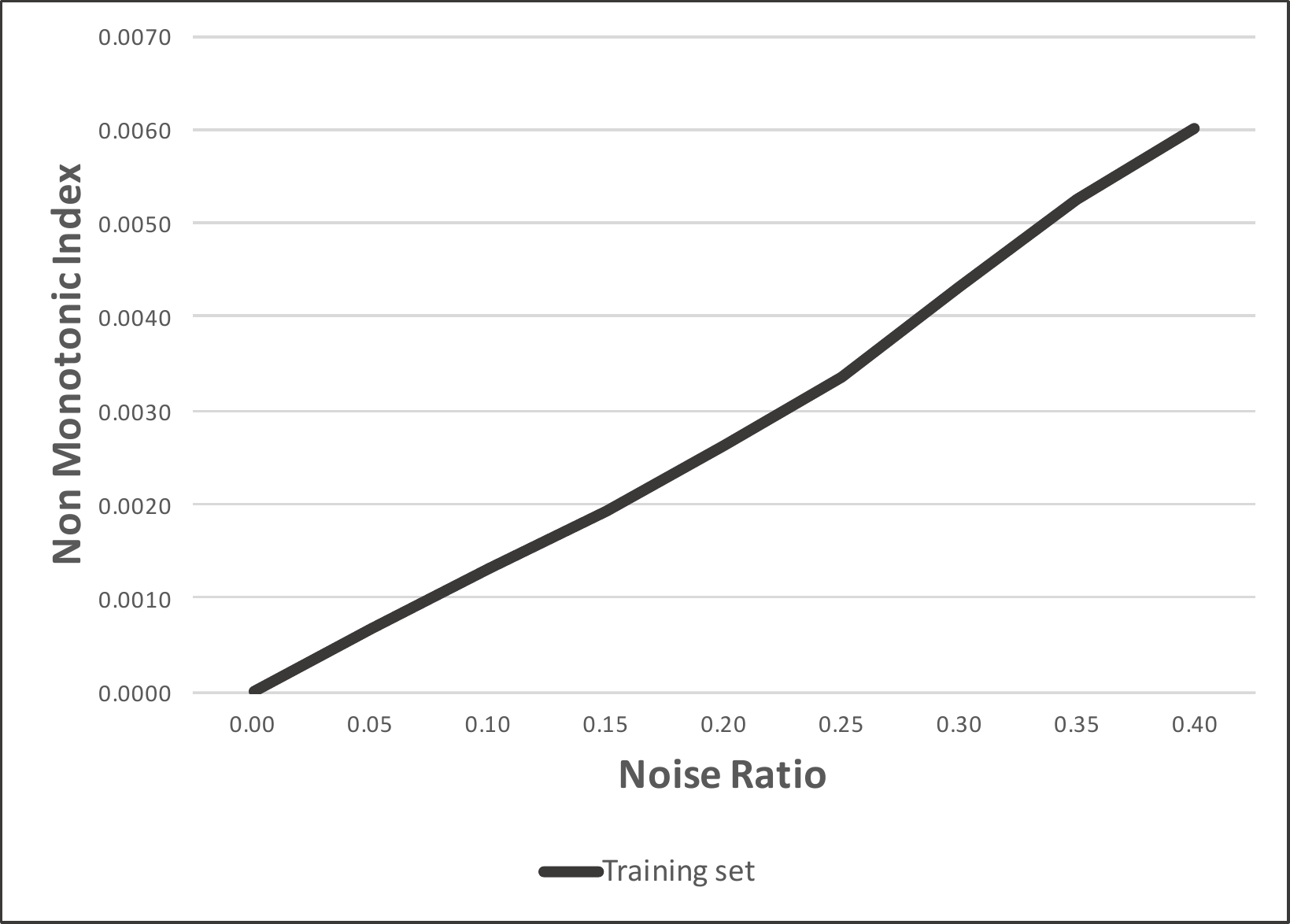}
 		\caption{Impact of the addition of class noise in Artiset on monotonic violations measured by NMI .}
		\label{fig:noiseData}		
	\end{center}
 \end{figure}

Figure \ref{fig:noiseData} shows the impact of increasing noise on the number of monotonic violations in Artiset training sets. This effect is measured by the Non-Monotonic Index (NMI) over the resulting training samples. As previously mentioned, class noise significantly aggravates the monotonicity of the data-sets. The increase in NMI is directly proportional to the increase in noise as clearly shown in Figure \ref{fig:noiseData}.

\begin{figure}[tb]
 	\begin{center}
     \subfloat[Noise effect in terms of Accuracy.\label{fig:noiseACC}]{\includegraphics[scale=0.35]{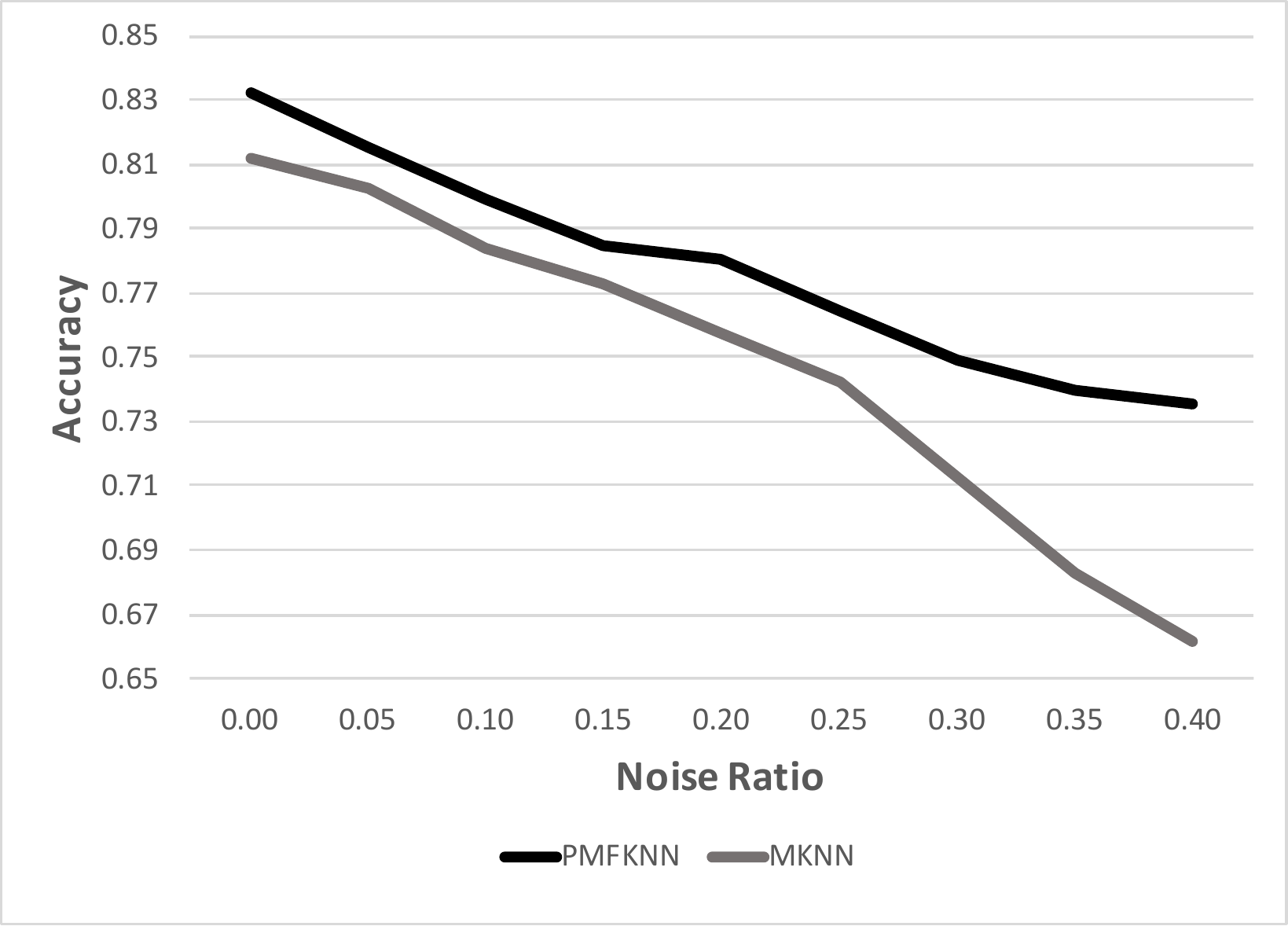}}
 	\hspace{0.5mm}
 	\subfloat[Noise effect in terms of MAE.\label{fig:noiseMAE}]{\includegraphics[scale=0.35]{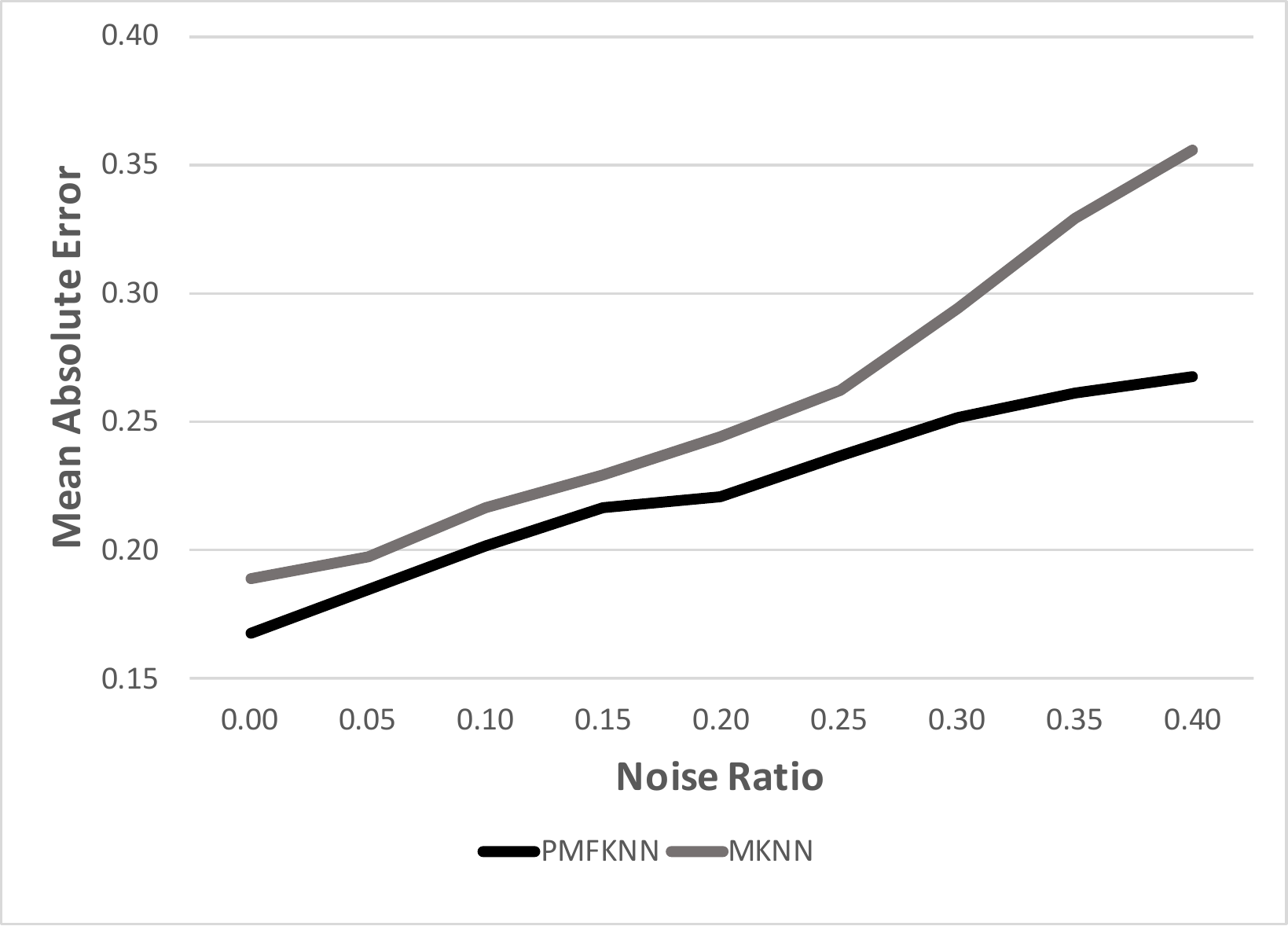}}
 	\hspace{0.5mm}
 	\subfloat[Noise effect in terms of NMI.\label{fig:noiseNMI}]{\includegraphics[scale=0.35]{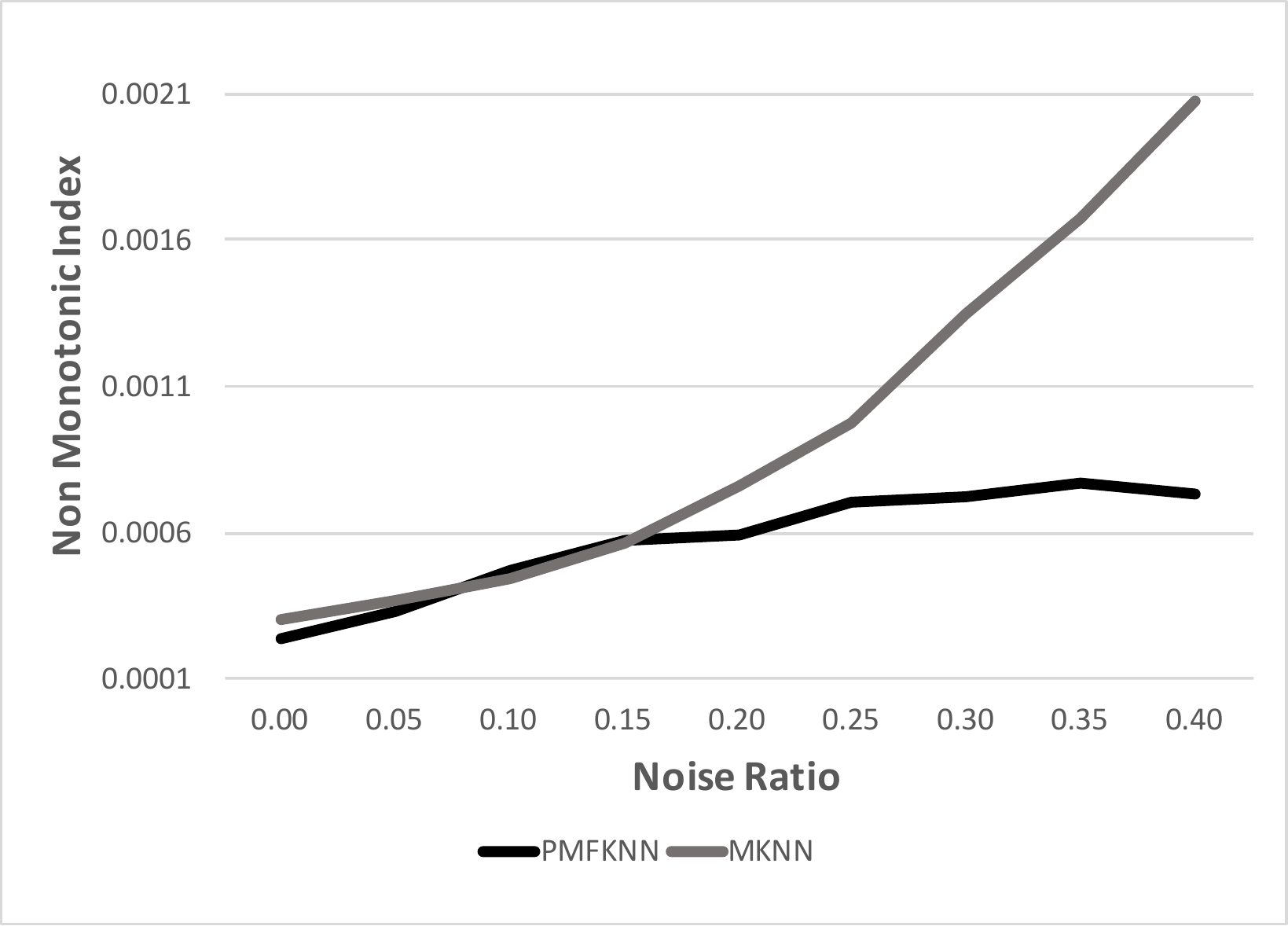}}
 		\caption{Comparison of MonF$k$NN-PM and M$k$NN performance on Artiset data-set with the different amounts of noisy samples.}
		\label{fig:noise1}		
	\end{center}
 \end{figure}

Figure \ref{fig:noise1} shows the performance of MonF$k$NN-PM and M$k$NN (darker and lighter lines, respectively) on the basis of precision (\ref{fig:noiseACC}), MAE (\ref{fig:noiseMAE}) and NMI (\ref{fig:noiseNMI}), with the progression of noise. As expected, while the amount of noise grows, the performance of both methods get worse, that is, their accuracy decreases and errors and non-monotonic predictions increases. However, there are some big differences between classifiers. 

Firstly, the behavior of MonF$k$NN-PM facing noise is clearly better than that of M$k$NN in every tested aspect. The black lines are always located above the lighter ones in Figure \ref{fig:noiseACC}, which indicates greater accuracy, and under them in Figures \ref{fig:noiseMAE} and \ref{fig:noiseNMI}, meaning better MAE and NMI for MonF$k$NN-PM. Usually, the distance between both methods is large, with the exception of the NMI results obtained for the smallest values of noise. In addition, while the noise ratio increases, their differences also increase.  

The slope of deterioration of MonF$k$NN-PM performance remains stable, even being reduced in some cases, while the M$k$NN slope becomes steeper as the amount of noise increases. This last event can be clearly seen when the noise ratio reaches the 25\% of the instances, where the decline of M$k$NN is magnified, especially in terms of monotonicity (Figure \ref{fig:noiseNMI}). On the other hand, the NMI results of MonF$k$NN-PM seem to increase at a slower rate by that point. This exhibits the great robustness of MonF$k$NN-PM to monotonic violations.

\begin{figure}
 	\begin{center}
     \subfloat[Exact decision surface for Artiset. Black points represent noise. \label{fig:borderReal}]{\includegraphics[trim={1.5cm 0.5cm 1.5cm 0.5cm},clip,scale=0.33]{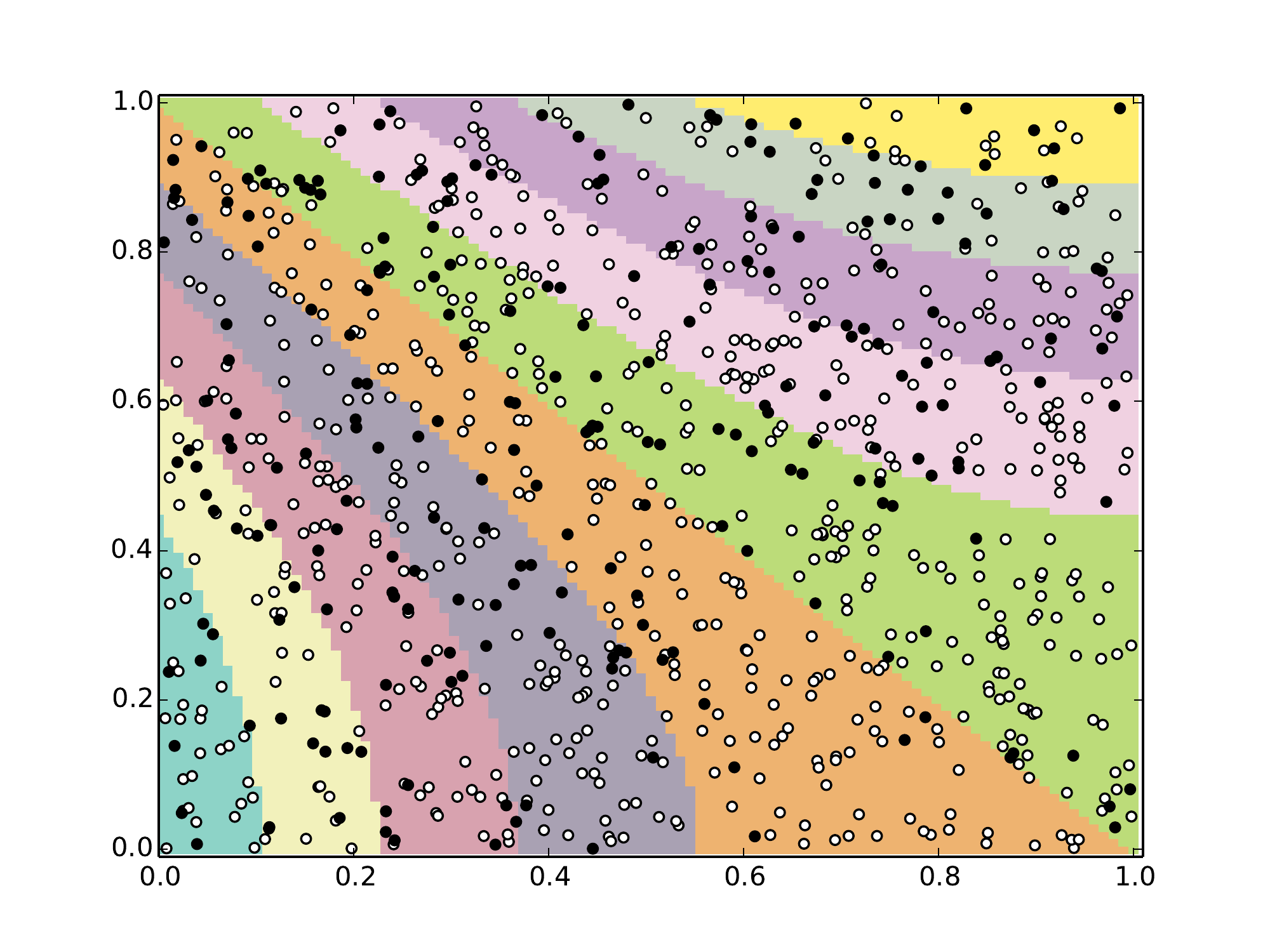}}
 	\hspace{0.5mm}
 	\subfloat[Decision surfaces inferred by M$k$NN.\label{fig:borderMKNN}]{\includegraphics[trim={1.5cm 0.5cm 1.5cm 0.5cm},clip,scale=0.33]{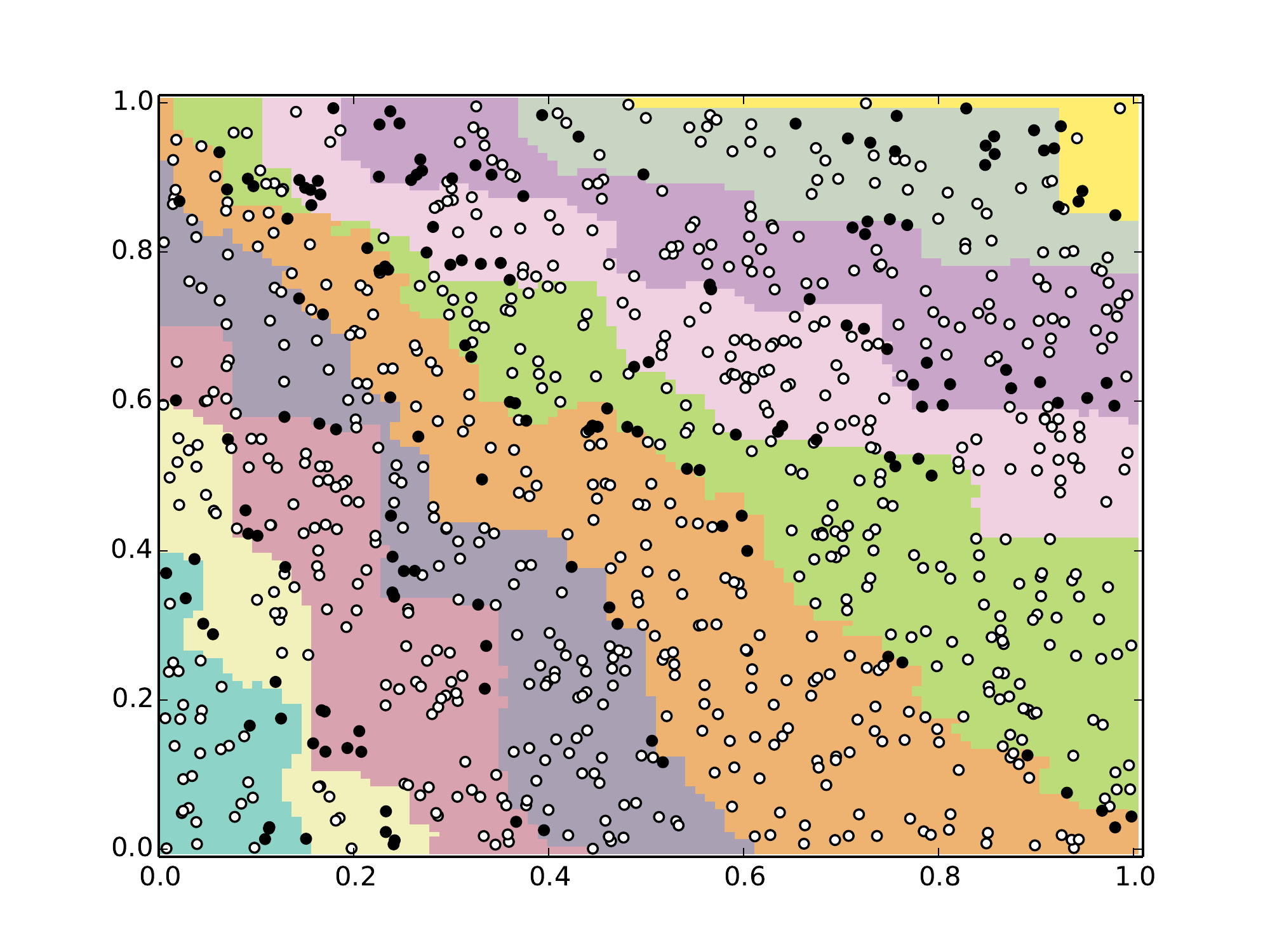}}
 	\hspace{0.5mm}
 	\subfloat[Decision surfaces inferred by MonF$k$NN-PM.\label{fig:borderMFKNN}]{\includegraphics[trim={1.5cm 0.5cm 1.5cm 0.5cm},clip,scale=0.33]{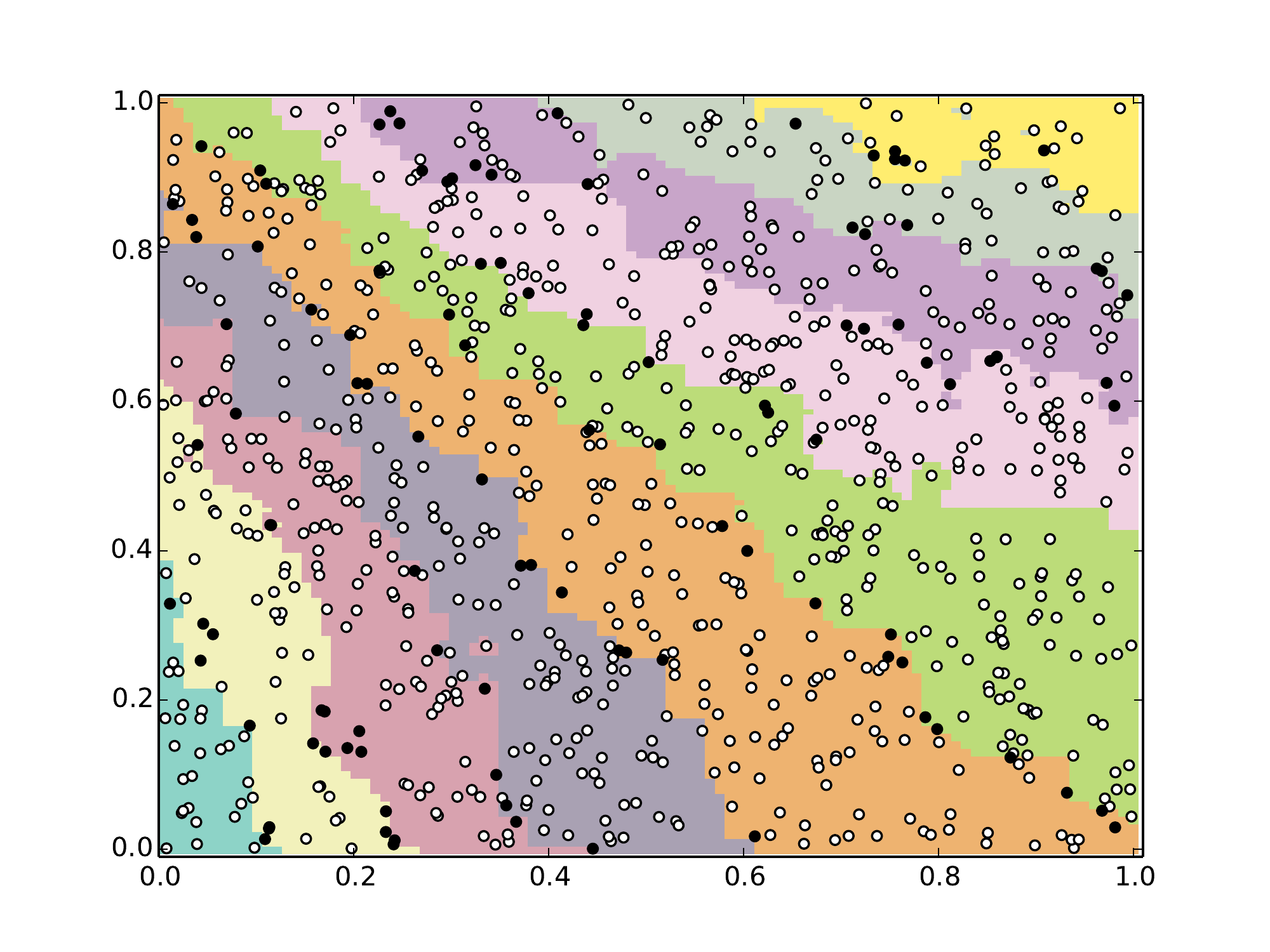}}
 		\caption{Classification boundaries inferred by M$k$NN and MonF$k$NN-PM from the plotted Artiset with 35\% noisy instances. Black points represent the instances wrongly classified by the decision surfaces shown.}
		\label{fig:noise2}		
	\end{center}
 \end{figure}

Next, the behavior of both methods in relation to noise are analyzed using a graphical example. Figure \ref{fig:noise2} is a graphical representation of the predictions and classification boundaries inferred by M$k$NN and MonF$k$NN-PM for \textit{Artiset} with 35\% noise. Figure \ref{fig:borderReal} represents the perfect class surfaces defined by \textit{Artiset} generation expression (see Section \ref{sec:framework}) and the training samples. In Figure \ref{fig:borderReal}, black points represent the noise artificially introduced into the data-set. In Figures \ref{fig:borderMKNN} and \ref{fig:borderMFKNN}, the black examples are wrongly classified instances, while the right predictions are colored in white.

The first clear difference between the M$k$NN and MonF$k$NN-PM performances shown in Figures \ref{fig:borderMKNN} and \ref{fig:borderMFKNN} is the amount of black dots. MonF$k$NN has far fewer classification mistakes than M$k$NN. Additionally, MonF$k$NN-PM is better at conserving the right regions for the classes, while M$k$NN can lose nearly all the entire sections of some of them. The regions in lighter and brighter yellow are shrunk by M$k$NN in favor of their adjacent classes.

With these experiments, MonF$k$NN has shown strong robustness to monotonic noise preserving the decision boundaries as precisely as possible, and hence, has performed well in terms of precision, error costs and monotonicity. This robustness is the result of all the procedures included in MonF$k$NN, but it may also be mainly due to the reduction of the impact of non-monotonic noise during the extraction of the class memberships of the training instances.  
 
\newpage

\section{Conclusion}
\label{sec:conclusions}

In this paper, we proposed a Fuzzy $k$-Nearest Neighbors model for classification with monotonic constraints. The final class label obtained from membership functions has been revised to respect these constraints. MonF$k$NN has been designed with different mechanisms to reduce the influence of monotonic violations. As a demonstration of its flexibility, two different model configurations with different behaviors have been presented.

Over the course of the experimental analyses, the great potential of both proposed versions, namely Pure and Approximate Monotonic Fuzzy $k$-NN, has been shown in relation to monotonicity and accuracy, respectively. Compared to other methods, MonF$k$NNN is significantly better in terms of accuracy and error cost, matching the best NMI results. In addition, it has shown its robustness to large amounts of noise while preserving its good performance. 



Future proposals should be robust to monotonic noise in order to obtain accurate and monotonic predictions. MonF$k$NN, as an example, opens possibilities to other fuzzy approaches since they are also potentially reliable against noise. Additionally, fuzzy techniques may be useful when defining different levels of constraints between input and output attributes. That is, some attributes may be more important than others regarding monotonicity. This problem representation may be very useful for monotonic classifiers.

\section*{Acknowledgements}
This work was supported by the Spanish Ministry of Economy and Competitiveness under Grant TIN2017-89517-P and a research scholarship (FPU) given to Sergio Gonz\'alez by the Spanish Ministry of Education, Culture and Sports.

\bibliographystyle{model1b-num-names}
\bibliography{Bibliografia}

\end{document}